\documentclass{ieeeaccess}
\pdfoutput=1 
\usepackage{cite}
\usepackage{amsmath,amssymb,amsfonts}
\usepackage{algorithmic}
\usepackage{graphicx}
\usepackage{textcomp}
\usepackage{subfig}
\usepackage{mathtools}
\usepackage{hyperref}
\graphicspath{{./Figures/}}

\newcommand{\argmin}{\operatornamewithlimits{argmin}}

\def\BibTeX{{\rm B\kern-.05em{\sc i\kern-.025em b}\kern-.08em
    T\kern-.1667em\lower.7ex\hbox{E}\kern-.125emX}}
\begin{document}
\history{Date of publication xxxx 00, 0000, date of current version xxxx 00, 0000.}
\doi{10.1109/ACCESS.2017.DOI}

\title{From noisy point clouds to complete ear shapes: unsupervised pipeline}

\author{\uppercase{Filipa Valdeira}\authorrefmark{1},
\uppercase{Ricardo Ferreira}\authorrefmark{2}, \uppercase{Alessandra Micheletti}\authorrefmark{3} and \uppercase{Cláudia Soares}\authorrefmark{4}}
\address[1]{Department of Environmental Science and Policy, Università degli Studi di Milano, via Celoria 2, 20133 Milan, Italy (e-mail: filipa.marreiros@unimi.it)}
\address[2]{$\mu$Roboptics, Lisbon, Portugal (e-mail: ricardo.ferreira@roboptics.pt )}
\address[3]{Department of Environmental Science and Policy, Università degli Studi di Milano, via Celoria 2, 20133 Milan, Italy (e-mail: alessandra.micheletti@unimi.it)}
\address[4]{ NOVA LINCS, Computer Science Department, NOVA School of Science and Technology, Universidade NOVA de Lisboa, 2829-516 Caparica, Portugal(e-mail: Claudia.soares@fct.unl.pt )}

\tfootnote{This work has received funding from the European Union’s Horizon 2020 research and innovation programme  under the Marie Skłodowska-Curie Project BIGMATH, Grant Agreement No 812912. This work was partially funded by the Eureka Eurostars project E!11439 FacePrint. }

\markboth
{Author \headeretal: Preparation of Papers for IEEE TRANSACTIONS and JOURNALS}
{Author \headeretal: Preparation of Papers for IEEE TRANSACTIONS and JOURNALS}

\corresp{Corresponding author: Filipa Valdeira (e-mail: filipa.marreiros@unimi.it).}

\begin{abstract}
Ears are a particularly difficult region of the human face to model, not only due to the non-rigid deformations existing between shapes but also to the challenges in processing the retrieved data. The first step towards obtaining a good model is to have complete scans in correspondence, but these usually present a higher amount of occlusions, noise and outliers when compared to most face regions, thus requiring a specific procedure. Therefore, we propose a complete pipeline taking as input unordered 3D point clouds with the aforementioned problems, and producing as output a dataset in correspondence, with completion of the missing data. We provide a comparison of several state-of-the-art registration and shape completion methods, concluding on the best choice for each of the steps. 
\end{abstract}


\titlepgskip=-15pt

\maketitle

\section{Introduction}
\label{intro}
Modelling shapes from 3D point clouds, and in particular human shapes, is useful in a wide range of applications and an ongoing topic of research. Here, we focus on the particular case of studying the variability of smaller detailed parts of the head, so that in the future their relationship with the remaining parts can be modelled. Therefore, the final goal is to obtain a model of these specific areas, that best describes their variation. Our driving example in this context will be the modelling of a ear and the description of how it fits the overall head.

Given the increased interest in human face models and availability of 3D data, this area has seen several developments in the recent years, leading to overall high quality full face and full head models \cite{article:3DMM_LYHM,article:3DMM_large_scale,article:Gaussian_Process_Combining_3DMM}. However, it is known that such models present poor quality or variability in more detailed parts, such as the ears \cite{article:Gaussian_Process_Combining_3DMM_Ear}. Reasons for such performance degradation are: the variability of the whole head shape surpasses the one of these smaller parts and data occlusion in these regions, since data are usually formed by 3D point clouds, obtained by scans of the head. There is a need for methods focusing on the modelling of these kinds of shapes, for example for prosthesis design, and this is where this work is positioned.

In order to obtain a model from a set of raw scans, there are two main steps that should be taken. The first is how to relate the scans amongst themselves so that each point has the same semantic meaning across all scans (registration). The second is how to find a way to describe the variability of shapes in a lower dimension (modelling). The two steps are related and the quality of one influences the other. 

The registration of head scans has several existing solutions providing good results for the most part of the shape and particularly for the face, since many methods focus only on this portion \cite{article:3DMM_Basel_Face_Model,article:Morphable_Blanz_Vetter,article:3DMM_large_scale}. However, as stated above, areas such as the ears are difficult to capture with the existing scanning methods, leading to several data problems, which make the registration more challenging and often cause the available solutions to fail or perform poorly. 

Naturally, the existence of previous models for those areas could help in the registration by providing prior information on the shapes. Yet, the existence of those models would require a previous registration. In fact, for the particular case of ear models, the existing work is still limited and often makes use of manual work to help in the registration task \cite{article:3DMM_Ears_Dai_Data_augmented_journal_citation,article:3DMM_Ears_Zolfaghari_LDDMM,article:Ear_model_template_based}.

Therefore, our main contributions are
\begin{itemize}
	\item comparison of state-of-the-art registration methods for the challenging case of 3D ear registration;
	\item a fully unsupervised pipeline that takes 3D ear data obtained from raw scans, with extensive data problems (missing data, outliers and noise) and produces registered and complete shapes \footnote{Code available at \url{https://github.com/FilVa/Shapes\_pipeline}};
	\item a new ear dataset obtained from real 3D scans of human heads.
\end{itemize}

\subsection{Related work}
\label{subsec:related_work}
Similar work consists of complete pipelines providing a path from 3D point clouds to shape models, in particular human shapes. First, we consider solutions for the human head and face, in general, and explain why they are not suitable for the particular case of the ear. We then focus on existent work on ear modelling and the current limitations.

\subsubsection{Face and head models}
A state-of-the-art approach for human face modelling are the 3D Morphable Models (3DMM). They are commonly obtained from raw scans by applying a registration method followed by Principal Component Analysis (PCA) , with possible intermediate steps for performance improvement. The first 3DMM was proposed in \cite{article:Morphable_Blanz_Vetter} and serves as basis for many subsequent approaches. Correspondence between scans is achieved by a gradient-based optic flow algorithm, after which PCA is used to obtain a low dimensional parametric model. An improved version was then achieved with the Basel Face Model (BFM) \cite{article:3DMM_Basel_Face_Model}, where dense correspondence is performed with optimal non-rigid Iterative Closest Point (NICP) step \cite{article:Registration_NICP_2007}, leading to better results. More recently, the Large Scale Facial Model (LSFM) was learnt from 10.000 examples, in \cite{article:3DMM_large_scale}. The dataset has the particular advantage of including a large diversity of age, gender and ethnicity on the training samples, leading to a more complete model. The pipeline consists of 3D landmark localisation, followed by NICP for dense correspondence, automatic detection and exclusion of failed correspondences and finally PCA.

While the previous models achieve very good results, they are limited to the face region and it is often useful or required to have a model of the entire head, as we seek in this work. Therefore, the first 3DMM of the full head, the Liverpool-York Head Model (LYHM), was proposed in \cite{article:3DMM_LYHM}. It starts with landmark localization in 2D and projection to 3D, followed by pose normalisation. Then, dense correspondence is achieved by a variation of the Coherent Point Drift (CPD) \cite{article:Registration_CPD}. This is followed by Generalized Procrustes Analysis (GPA) and PCA. However, it is noted that this model still lacks precision in some regions, since the variance of cranial and neck areas dominate over the face in the PCA parametrization. 

In order to overcome this challenge, the authors in \cite{article:Gaussian_Process_Combining_3DMM} approach the topic of combining different 3DMM, in order to capture the different benefits of each model. In this case, they merge the LSFM due to its great representation of facial detail and the LYHM as it represents the full head. While they are able to capture the variability of the face within a full head, the model still lacks expressibility for the detailed regions, such as ears or eyes. 

Consequently, the latter method was extended in \cite{article:Gaussian_Process_Combining_3DMM_Ear} to overcome this shortcoming. This is the work which is more closely related to our contribution, in the sense that it provides a model able to relate the ear with the entire head shape. However, it is noted that the ear model requires the identification of 50 manual landmarks for registration. Besides, we approach the problem in a different way, aiming at first reconstructing the problematic ear data in the raw scans and then building the full model, while in \cite{article:Gaussian_Process_Combining_3DMM_Ear}, the head model is augmented with an ear model.

\subsubsection{Ear models and dataset}
Given the previous overview, it is evident that obtaining detailed representations of ears from raw scans of the entire head directly is not a straightforward task and, in principle, requires some previous model of such region. Therefore, it is pertinent to get a grasp on the existing work on ear models. To the extent of our knowledge there are three proposed models of the ear in \cite{article:3DMM_Ears_Zolfaghari_LDDMM}, \cite{article:3DMM_Ears_Dai_Data_augmented_journal_citation} and \cite{article:Ear_model_template_based}. However, only the second final model, that is the mean shape and principal components, is made publicly available.

The work in \cite{article:3DMM_Ears_Zolfaghari_LDDMM} makes use of the large deformation diffeomorphic metric mapping (LDDMM) framework to produce the model. Under this setting, the authors model the deformation of one shape into the other as a flow of diffeomorphisms. This model is then simplified with a kernel based Principal Component Analysis, thus obtaining a morphable model. To build the model they use the SYMARE database \cite{database:SYMARE}, composed of 58 ears. They evaluate the model by computing using 57 samples as training, and testing the accuracy of shape reconstruction of the left out ear.

In \cite{article:3DMM_Ears_Dai_Data_augmented_journal_citation}, the authors start with a limited dataset of 20 samples from 10 subjects, building an initial model with CPD and PCA.  However, given its reduced variability, the authors propose to use the initial model along with an existing 2D ear database already labelled with landmarks to produce a larger augmented dataset. First, they find the parameters of the model leading to a shape that, when projected to 2D, is the most similar to the 2D image in the dataset. Then, they deform the mean shape of the initial model to match each of the ears, with a variation of CPD. Therefore, they obtain a final dataset of 600 ears, all with the same number of points and in correspondence. The final model is obtained from this dataset in a straightforward manner with GPA and PCA. The model is incrementally improved by iterating over the data augmentation and model production step.

In \cite{article:Ear_model_template_based}, the authors propose a full pipeline from ear color images to a complete 3D model, consisting of three main steps. They start with detection of the ear and landmarks, using CNN networks, which provide an initial recognition of the ear object in the image. This is followed by two steps of registration: rigid one with ICP and non-rigid one with non-rigid ICP. The first aims at removing translation and rotation, while the second deals with shape differences. The experimental results after the full pipeline are compared visually with the 2D images of the ears.

The field of 3D ear modelling, albeit the interesting and deep work already developed, is far from achieving the necessary accuracy in applications. We are able to conclude that current methods typically require some manual annotation to start the pipeline. Furthermore, there is only one publicly available 3DMM model. Although it is a good starting point, it is composed of very few real scans, since the majority of the dataset was created by sampling of a fitted generative model, which can bias the learned model. Therefore, we propose to retrieve the ear data from raw full head scans. Naturally, identifying which points belong to the ear is not a straightforward task, mostly due to the scan problems existing in these regions. So, the main obstacle is to find a registration method able to overcome these challenges.

\subsection{Paper Outline}

In Section~\ref{sec:prob_form} we start with a clear formulation of the problem at hand, a description of the proposed pipeline to overcome each obstacle and a justification for our choice of dataset and metrics. Then, in Section~\ref{sec:zoom_pipeline_steps}, we cover different possible solutions for each step of the pipeline, providing the theoretical background for the methods to be compared afterwards. Section~\ref{sec:results} contains the numerical results, where for each pipeline step the performance of different methods is extensively evaluated. From that, we conclude on the best choice of method sequence, presented in Section~\ref{sec:concl}, together with some final remarks and future work.

\section{Setting and pipeline overview}
\label{sec:prob_form}

\subsection{Problem formulation and proposed pipeline}

\begin{figure}[htp]
	\subfloat[Raw full head scan ]{\label{fig:data_full_head}
		\includegraphics[width=.2\textwidth,clip, trim={100pt 330pt 160pt 100pt}]{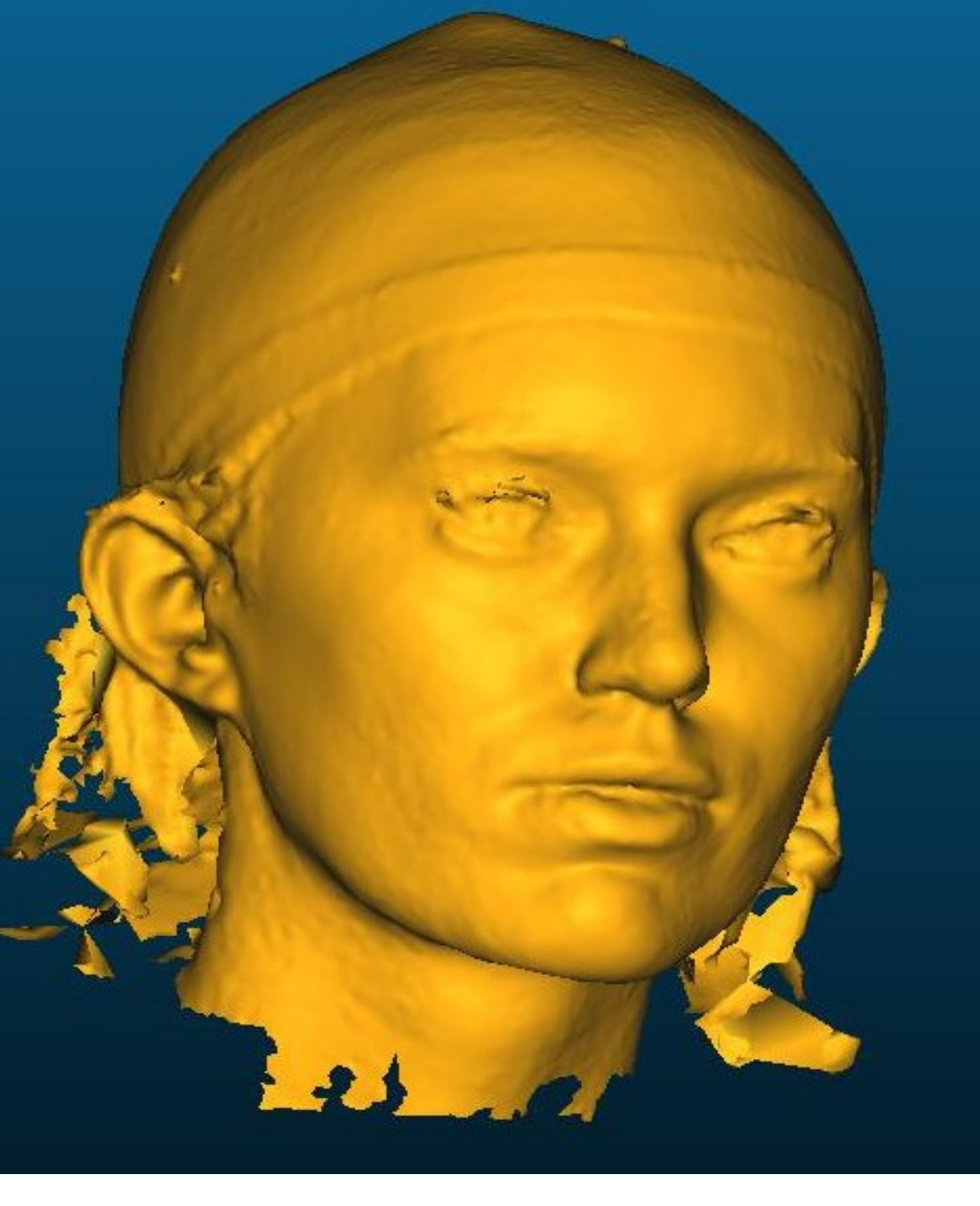}}
	~
	\subfloat[Retrieved ear part]{\label{fig:data_raw_ear}
		\includegraphics[width=.2\textwidth,clip, trim={80pt 170pt 150pt 220pt}]{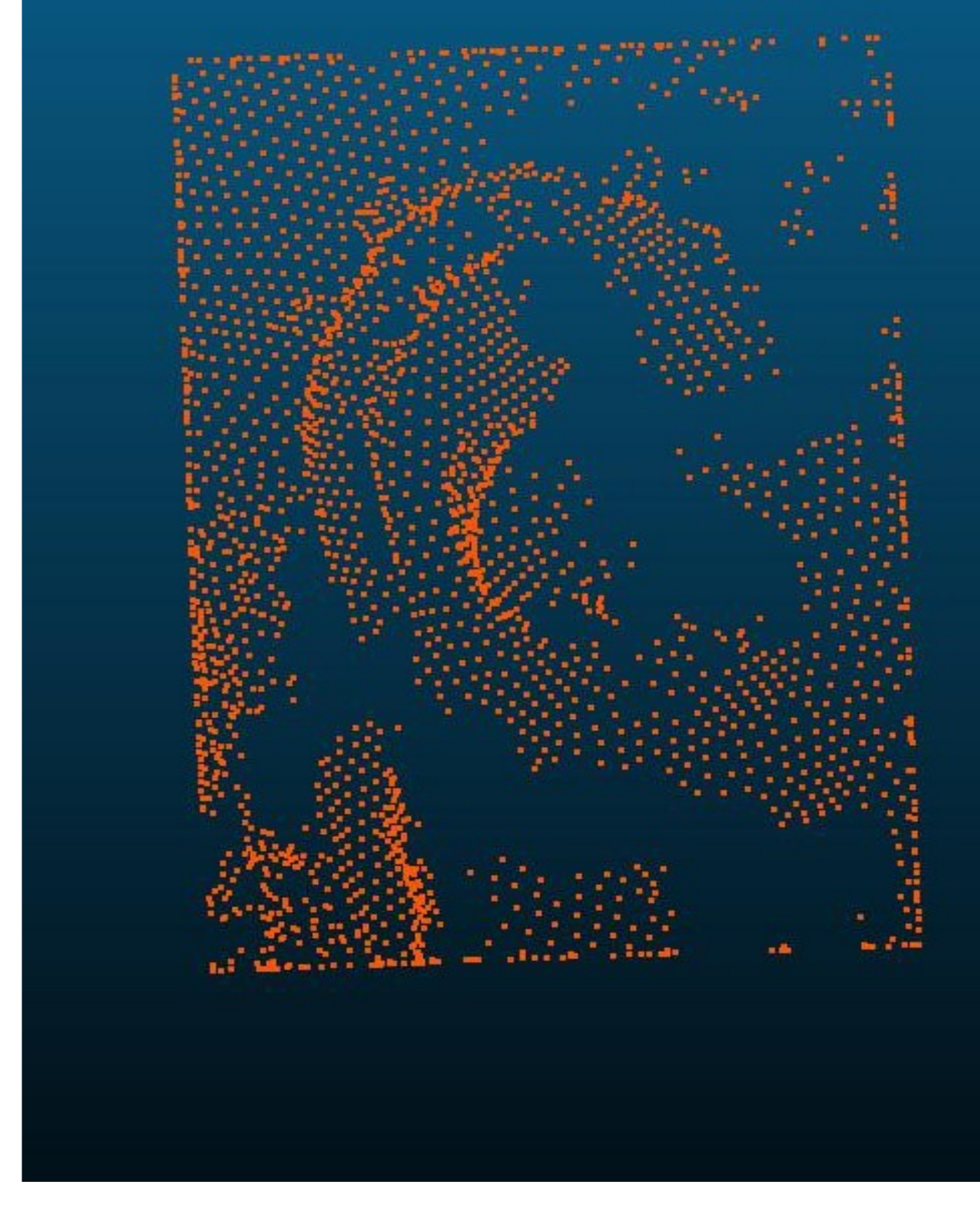}}
	
	\caption{Original data from  dataset \cite{article:3DMM_LYHM}}
\end{figure}

The final goal of this work is to relate the parameters of a ear model with the shape of a full head. Therefore, we start with a publicly available dataset of full human heads found in \cite{article:3DMM_LYHM}, composed of scans such as the one in Fig.~\ref{fig:data_full_head}. From now on, we will call this the Head dataset. It is immediately evident that the ear region, because of its shape complexity, presents a considerable amount of data problems not found in other sections. Besides, as these are point clouds, they are not in correspondence, and so this must be the first step in producing a model.

\begin{figure}
	\centering
	\includegraphics[scale=0.6,clip, trim={50pt 0pt 0pt 0pt}]{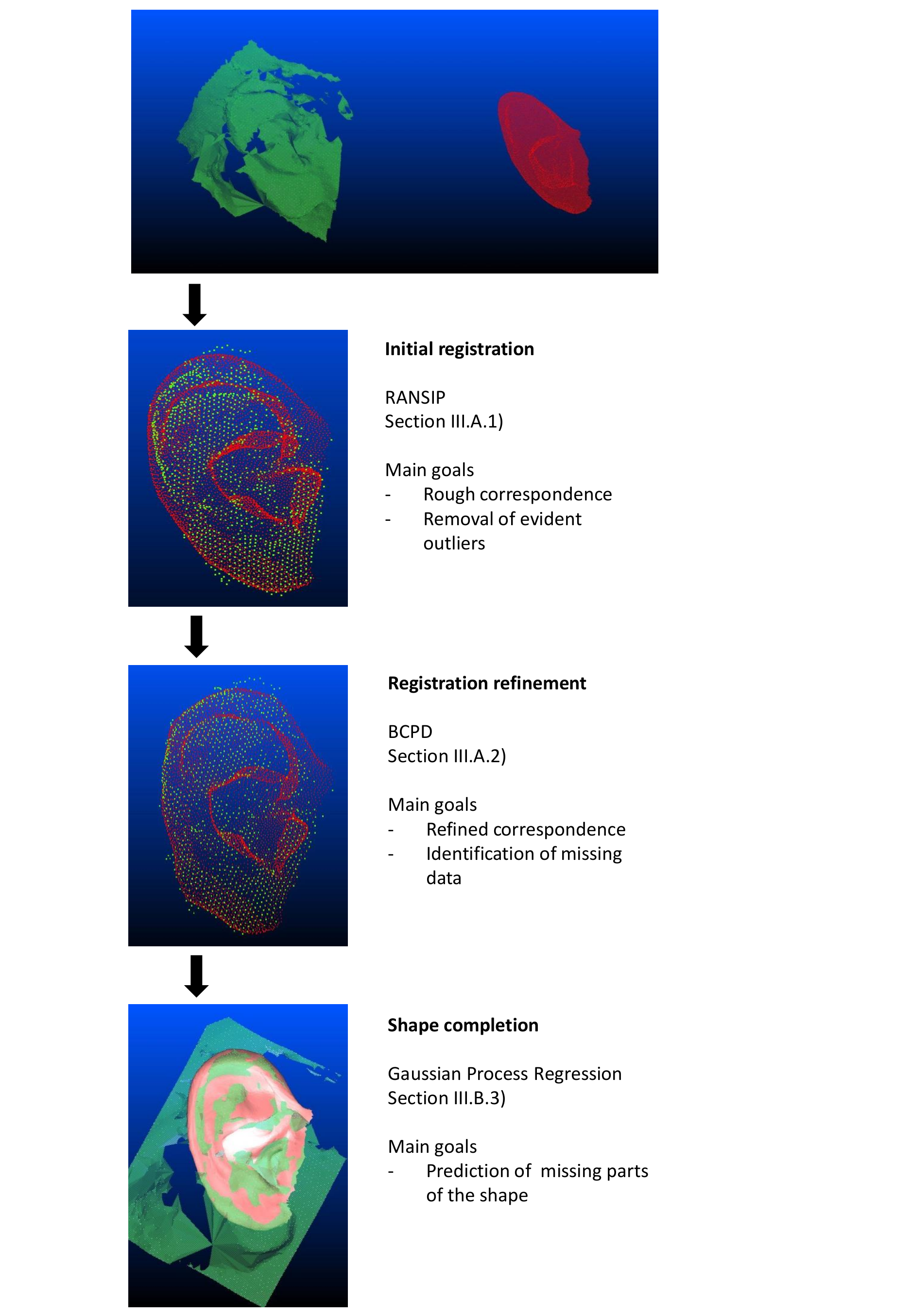}
	\caption{Pipeline scheme. In red we observe the template being registered to the target shape in green. The top image shows the initial position of both point clouds, where we can see that there is considerable rotation and translation between them. The first step is the initial registration which should place both shapes in a reasonable aligment and remove the most clear outliers, in general the points of the skin around the ear. The result of this step is depicted in the second image, where the green points are the inliers kept from the initial target. Afterwards an improved correspondence is achieved by allowing no-rigid deformations with BCPD. Finally, the points of the template without correspondence are predicted with Gaussian Process Regression, producing the red shape in the final picture, thus originating the Reconstructed Dataset. }
	\label{fig:pipeline_scheme}
\end{figure}

As a first approach, we would apply an existing registration method to the full head or separately to the ear, obtain a model with a dimensionality reduction approach such as PCA and study the variation of parameters of the ear model with the head model. However, the methods used for head registration tend to fail on the ears, thus calling for an alternative approach. We propose the pipeline in Fig.~\ref{fig:pipeline_scheme}.

The first step is merely a cut of the ear region from each head scan, resulting in samples such as the one in Fig.~\ref{fig:data_raw_ear}. Since the head scans are already aligned (and this can always be done with several existing methods), we only need to extract the same region for every scan in an unsupervised manner. 

\begin{figure}[htp]
	\subfloat[Mean shape of the Ear Dataset \cite{article:3DMM_Ears_Dai_Data_augmented_journal_citation} ]{\label{fig:mean_shape}
		\includegraphics[width=.21\textwidth,clip, trim={50pt 330pt 150pt 80pt}]{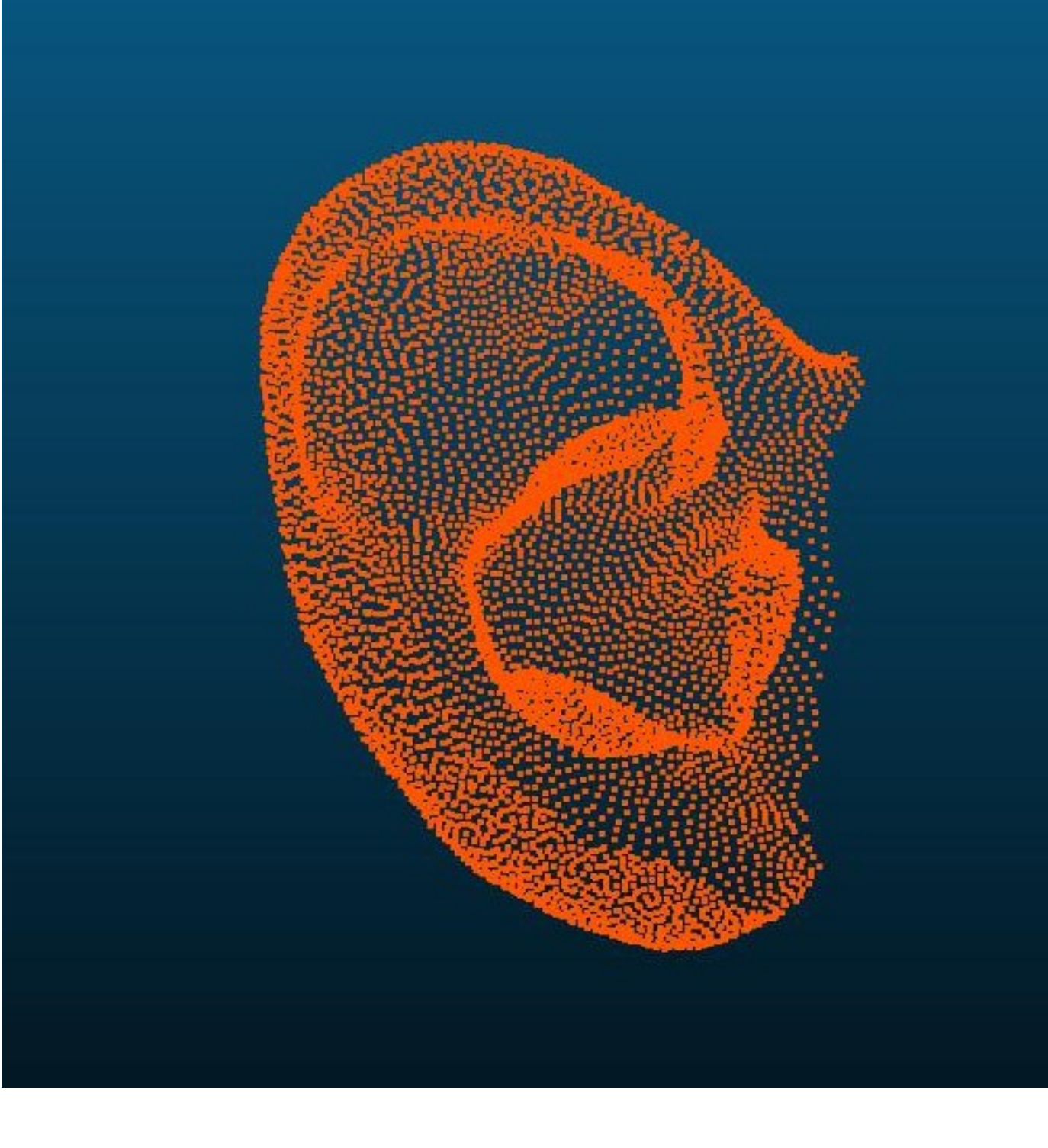}}
	~
	\subfloat[Shape from the Simulated Dataset]{\label{fig:data_simulated}
		\includegraphics[width=.22\textwidth,clip, trim={50pt 350pt 150pt 80pt}]{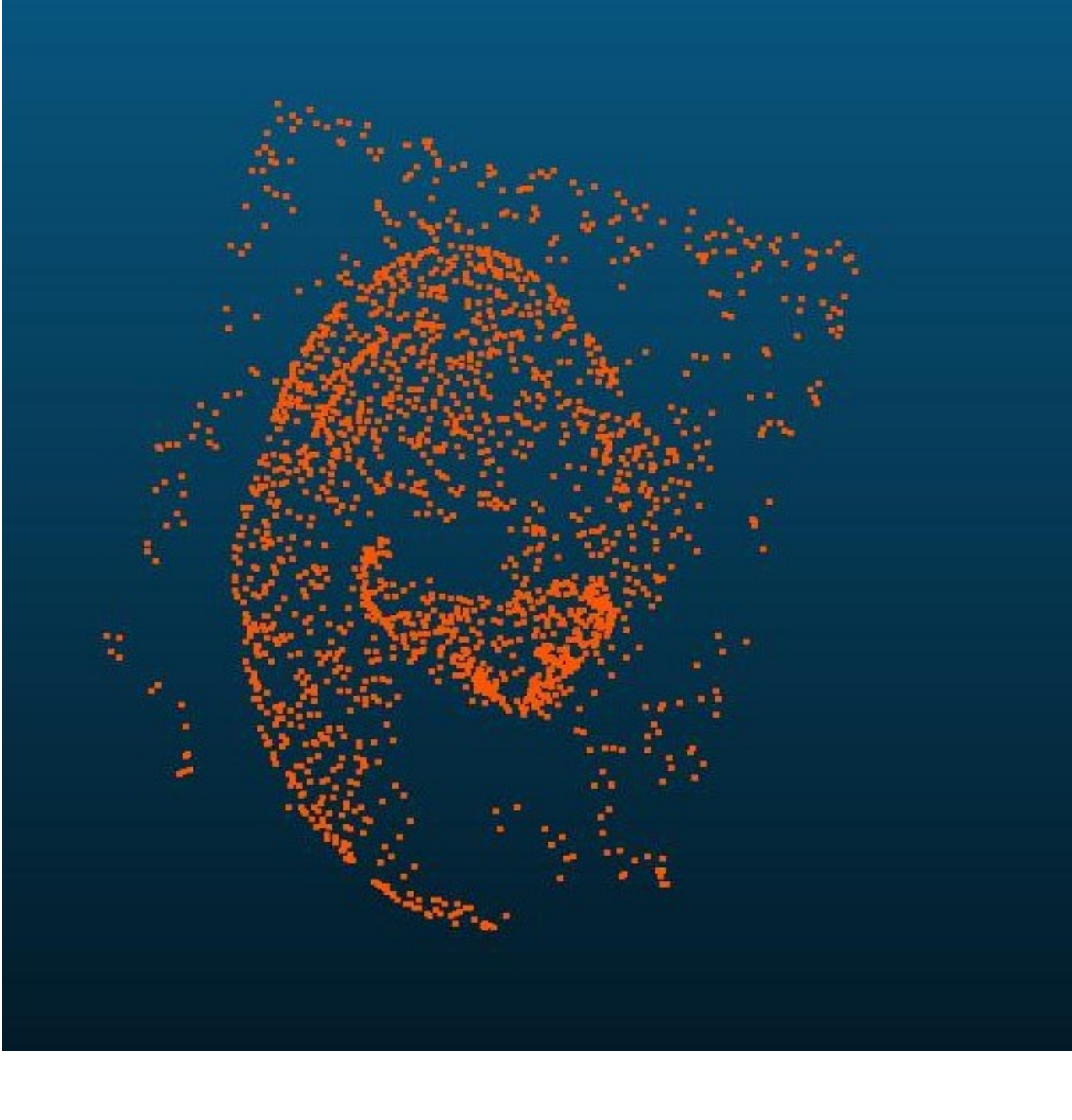}}
	
	\caption{Ear datasets}\label{fig:ear_dataset}
\end{figure}

To register the samples amongst themselves, we need a template, which should be a good representation of the ear shape without any missing data or  outliers. Therefore, we use the mean shape of the ear model available from \cite{article:3DMM_Ears_Dai_Data_augmented_journal_citation}, composed of 7111 points and represented in Fig.~\ref{fig:mean_shape}. 

The problem is then to register the template from Fig.~\ref{fig:mean_shape} with samples similar to Fig.~\ref{fig:data_raw_ear}. However, in order to choose the best method, we need to compute evaluation metrics and so we require a ground truth, that is, we need to know the correspondence a priori. This is evidently not the case of Fig.~\ref{fig:data_raw_ear}, as this is precisely the problem we are trying to solve, motivating us to use the entire dataset from \cite{article:3DMM_Ears_Dai_Data_augmented_journal_citation} (denoted as the Ear dataset) for testing purposes, where all the shapes already have the same number of vertices and are in correspondence. The obstacle is that there are no missing points, outliers or noise in any of the samples, while registration of real head scans will entail all these problems. The solution is then to introduce all these problems in the original data, in order to replicate as much as possible the difficulty of real scans, while still having the knowledge of the true correspondences. We explain this process in the following subsection.

As seen in the scheme, we propose two different registration steps. The first should place the template in an appropriate position with respect to the target, mostly in terms of rotation and translation, and should identify the most clear outliers. This is the case of the skin area existing around the ear. After this, it is expected that non-rigid methods are able to perform better, thus refining the correspondences between the template and target. Finally, given the high occurrence of missing data, we introduce a step for shape completion, where we try to obtain the deformed template that most resembles each target.

As a final output, we obtain the Reconstructed dataset, that is, the reconstruction of each ear of the Head dataset, after the application of the full pipeline.

\subsection{Datasets}
\label{subsec:setting_data}
\subsubsection{Missing data}
The obtained scans of the Head dataset have missing points, not only uniformly spread, but also concentrated in particular regions of the ear which are more difficult to capture by the scanning process. Therefore, in the Ear dataset we introduce
\begin{itemize}
	\item Uniform missing data: by randomly removing data points from the entire shape, with a ratio of 0.2, i.e. we uniformly remove 20\% of points in the whole shape.
	\item Structured missing data: by randomly removing data points from a particular area, with a higher ratio of 0.8. This consists of a single region, manually defined on the inner and bottom part of the ear and identified by observation of the Head Dataset, aiming at replicating it as close as possible. 
\end{itemize}

\subsubsection{Outliers}
The ear region of the Head dataset also contains outliers (points with no correspondence in the template), both uniformly spread and in a structured manner. In particular, the structured outliers come from the fact that when we cut the ear portion from the entire head of the scan we do not know exactly which points belong to the ear, and consequently include some extra points. Therefore, in the Ear dataset we add
\begin{itemize}
	\item Uniform outliers: by introducing additional points over a bounding box containing the entire shape, with a ratio of 0.1.
	\item Structured outliers: by introducing additional points over a particular area of the shape, with a higher ratio of 0.4. This regions were manually defined in order to cover the skin around the ear found in most scans, as observed in the example shape of Figure~\ref{fig:ear_dataset}.
\end{itemize}

\subsubsection{Measurement Noise}
For each point in the Ear dataset we introduce Gaussian noise with zero mean and a chosen variance, so that they are slightly displaced, to simulate the lack of complete accuracy in the screening process.

\subsubsection{Simulated dataset}
Altering the original data of the Ear dataset with the previously stated processes, we obtain a noisy dataset (denoted as Simulated dataset) that should replicate the point clouds obtained from the real head scans. Fig. \ref{fig:ear_dataset} exemplifies this process, showing the mean shape of the ear model (left) and the tampered ear produced from the original data (right), attempting to replicate the problems.

\subsection{Metrics}
\label{subsec:setting_metrics}
For comparison of the different registration methods we consider several metrics. 

\subsubsection{Metrics for registration focusing on measurement noise}

We consider two different metrics to evaluate the registration results :
\begin{itemize}
	\item Fraction of correspondences : computed as the number of correspondences found over the total number of points in the template for which a correspondence exists. We want this to be as close to 1 as possible. A common metric for similar cases would be the Jaccard Index, where given two sets ($A$ and $B$) we compute the similarity between them as $\frac{|A\cap B|}{|A\cup B|}$. In our case, if $A$ is the set of points of the template and $B$ the set of the target, our metric could be expressed in this notation as $\frac{|A\cap B|}{|A|}$. We use the latter since the template corresponds to the ground truth and we want to relate all samples to it, regardless of the number of points of each target.
	\item Distance error : for each point in the template we identify the true correspondence in the target and the registered point by the method, computing the Euclidean distance between them. The true correspondence is possible to retrieve since the original Ear dataset is registered.  This metric then expresses the average distance for all points in the shape. So, given the template point cloud with $M$ points, the original target point corresponding to the $m$-th point given as $x_m \in \mathcal{X}\subset \mathbb{R}$ and the registered target point given as  $\tilde{x}_m \in \tilde{\mathcal{X}}\subset \mathbb{R}$, this error is expressed as
	\begin{equation*}
	D = \frac{1}{M}\sum_{m=1}^M \|x_m-\tilde{x}_m \|.
	\end{equation*}
\end{itemize}

The two previous metrics must be evaluated simultaneously since we want to identify as many points of the template as possible (fraction of correspondences), but we also require them to be correctly matched (distance error).

\subsubsection{Metrics for registration focusing on outliers and missing data}
We need to specifically evaluate the performance of the methods when it comes to identify outliers and missing data, given that these are the two main challenges in our data. For this purpose, we will use metrics typically used in classification problems, since we are essentially classifying each point as either inlier, outlier or missing.

Therefore, taking as example the outliers, we define
\begin{itemize}
	\item True Positives (TP) as the outliers identified as such;
	\item True Negatives (TN) as the non outliers identified as such;
	\item False Positives (FP) as the points incorrectly classified as outliers;
	\item False Negative (FN) as the outliers not identified as such.
\end{itemize}
Considering this, we define the following metrics
\begin{itemize}
	\item Precision (or Specificity) = $\frac{TN}{TN+FP}$ (expresses the ability to correctly identify non outlier);
	\item Recall (or Sensitivity) = $\frac{TP}{TP+FN}$ (expresses the ability to correctly identify outliers).
\end{itemize}

The same reasoning can be applied to the missing data, by taking the metrics with the missing points instead of the outlier ones.

\subsubsection{Shape completion}

In this step, we want to measure how close a predicted shape is from the ground truth (original shape from the Ear dataset). For this purpose, we consider a reconstruction error taking the distance between each predicted points and the true one, averaging over the shape and the dataset.
\section{Zooming in on the pipeline steps}
\label{sec:zoom_pipeline_steps}

\subsection{Registration}

In order to compare different shapes it is necessary to have all instances in the same vector space. However, most 3D data retrieval produces point cloud data, which are unordered vectors and so it is necessary to employ a registration method to achieve correspondence between examples. 

In general, we can look for sparse or dense correspondence. The first matches a reduced amount of points through landmarks (distinctive features compared to their local context, such as the tip of the nose, corner of the eyes or mouth), while the latter matches a large number of points with similar topological meaning. Although more challenging to compute, dense correspondences are able to express more detailed structures, which is a requirement for us.

Typically, the registration problem involves transforming one shape (template) as close as possible into the other (target). The methods mostly differ on the kind of transformation they consider and what is defined as being close to. 

Here, we denote by X the target shape, corresponding to the scan data in this case, and by Y the template shape, corresponding to the mean shape from the Ear dataset. Both X and Y are point sets, with $X = (x_1, ... ,x_N)^T$ and $Y = (y_1, ... ,y_M)^T$, where $N$ and $M$ are the respective number of points in each point set. The problem is then to find a transformation $\mathcal{T} (Y,\theta)$ with unknown parameters $\theta$, bringing $Y$ as close as possible to $X$ according to some defined metric.

This section goes over the currently used registration methods, from which the most promising will be selected for comparison with our data. They are split under the three main areas usually considered : Iterative Closest Point variants, probabilistic approaches and graph matching.

\subsubsection{Iterative Closest Point (ICP) and variants}
One of the most popular registration method class is the Iterative Closest Point (ICP) \cite{article:Registration_ICP}. It considers only rigid transformations for $\mathcal{T}$, that is rotations $R$ and translations $t$. The goal is to find $R$ and $t$ minimizing the distance between each point in $Y$ and its closest point in $X$. That is, solving the optimization problem
\begin{equation*}
\min_{R\in \mathcal{SO}(3), t\in \mathbb{R}^3} \sum_i^{N_s} \rho(\|Ry_i+t-x_j \|) ,
\end{equation*}
where $x_j$ is the target point closest to the transformed source ($x_j = \argmin_{x_k \in X} \|Ry_i+t-x_k\|$) and $\rho$ is a robust loss function.

ICP solves this problem by iteratively finding the closest points of the template on the target and then finding the least-squares rigid transformation between such pairs of points. Next, a new correspondence is found and a new transformation computed, and so on \cite{article:Registration_ICP_Statistical}. It is still considered state-of-the-art, given its reduced time complexity and accuracy in plenty of cases. However, it requires the initial position of the two point sets to be moderately close, as it is quite sensitive to initialization. 

Further variants can be found, such as Go-ICP \cite{article:GO_ICP} where the authors try to find the optimal transformation through a branch and bound strategy, to avoid the local optima obstacle. Another line of work to solve this drawback are randomized strategies based on the RANSAC. A recent  and state-of-the-art example is the SDRSAC \cite{article:SDRSAC}, combining the randomized approach with graph matching. The idea of SDRSAC is to perform several iterations where we subsample both point clouds and run a matching step (SDR matching), thus obtaining a rigid transformation to deform the template into the target. After this, we can compute the number of non-outliers found, denoted as consensus. The output is the transformation resulting in the highest consensus, after a number of iterations defined by the current probability of finding an inlier correspondence. The SDR matching step takes $N$ points of two point clouds and finds the best correspondence of only part of the points. Here the best correspondence relates to how well the pairs of points in each cloud match each other, that is, how similar is the distance between a pair of points in cloud A to a pair in cloud B).

We propose an additional method (RANSIP) in the line of RANSAC with ICP, but taking into account the point cloud normals. We run several iterations of ICP, taking as initialization the vector going from one centre of mass to the other for the translation and a random rotation matrix. Given the ICP result we compute the correspondences between shapes and determine the inliers. Then, for each inlier vertex of the template and target we obtain the normals and, subsequently, the angle between the normals of corresponding points. The final cost is taken as the median of those angles, which should be as low as possible. Similarly to SDRSAC, the number of iterations to run is defined by the probability of finding an inlier correspondence, but we only consider inliers with normals differing by less than  $\frac{\pi}{4}$ rad.

Furthermore, there is a non-rigid version of the ICP, the optimal step non-rigid ICP (NICP) proposed in \cite{article:Registration_NICP_2007}. Based on the ICP, it assumes that each point of the template may undergo a local affine transformation represented by a $3\times 4$ transformation matrix $T_i$, so that the parameters are in a $4N \times 3$ matrix $T\coloneqq \left[T_1 ... T_N\right]$. Then, the parameters are found by minimizing the cost function defined as
\begin{equation}
E(T)\coloneqq E_d(T) + \alpha E_s(T) + \beta E_l(T).
\label{eq:nicp}
\end{equation}
The first term minimizes the distance between the deformed template and the target, including a weighting term for each point which is set to zero for no correspondence and 1 otherwise. The second term is a regularization on the deformation, so to limit the deformations to acceptable shapes. It penalises the weighted difference of the transformations of neighbouring vertices, with a parameter balancing the rotational and skew transformations against translation. The final term guides the registration with landmarks. The NICP takes a sequence of decreasing $\alpha$ values and for each one iteratively finds the current correspondences and computes the deformations according to the cost function in~\eqref{eq:nicp}. The algorithm stops when consecutive transformations are similar enough.

\subsubsection{Probabilistic approaches}
The most used and representative method under this category is the Coherent Point Drift (CPD) \cite{article:Registration_CPD} considering the alignment of two sets as a probability density estimation problem.  This approach takes $Y$ (the template) as a set of centroids coming from a Gaussian Mixture Model (GMM) and $X$ (the target) as points generated by the centroids. The transformation $\mathcal{T}$ can be set as rigid, affine or non-rigid, producing 3 different versions of CPD. An important detail is that the centroids are forced to move coherently as a group, thus preserving the topological structure of the points (motion coherence constraint over the velocity field). The goal is to find the most likely centroid from which each point in $X$ was generated, thus resulting in a correspondence output. The GMM probability density function for this scenario is given as
\begin{equation*}
p(x) = \sum_{m=1}^{M+1} P(m)p(x|m),
\end{equation*}
where $p(x|m) = \frac{1}{(2\pi\sigma^2)^{D/2}} \exp{-\frac{\|x-y_m\|^2}{2\sigma^2}}$. It is then assumed that the probability of each point x belonging to centroid m is equal, so $ P(m)$ is set to $1/M$. Besides, an additional uniform distribution is added to account for noise and outliers $p(x|M+1) = 1/N$. This leads to the final pdf as 
\begin{equation*}
p(x) = w\frac{1}{N} + (1-w)\sum_{m=1}^M \frac{1}{M}p(x|m),
\end{equation*}
where $w \in (0,1)$.

The general idea is to find the parameters for the transformation of Y so as to minimize the difference between the two shapes. For this, the authors use the Expectation Minimization (EM) algorithm. Finally, for each point x we can compute the posterior probability of having been generated by centroid m given as $P(m|x_n) = P(m)p(x_n|m)/p(x_n)$.

For the non-rigid case, the authors define the transformation as 
$\mathcal{T}(y_m) = y_m + v(y_m)$, where $v(y_m)$ is a displacement vector. CPD has two parameters controlling the amount of non-rigid deformation, $\lambda$ and $\beta$, related to the smoothness of the displacement field.

While still considered state-of-the-art, CPD does not handle particularly well outliers, missing data and different number of points between both point clouds. Consequently, variants of CPD have been developed in recent years to attempt to deal with such drawback \cite{article:CPD_Extended_correspondence_priors,article:CPD_unified,article:CPD_PreservingGlobal_LocalStructures,article:CPD_adaptive_template_Dai_Pears,article:CPD_membership}.
While in the original CPD membership probabilities are all equal, in \cite{article:CPD_PreservingGlobal_LocalStructures} the authors propose to assign them differently in order to encourage matching of points with similar local structure. This method is therefore more robust to deformation, outliers, noise, rotation and occlusion. A more recent version was proposed in \cite{article:Registration_CPD_LocalConnectivity}, also enforcing the preservation of local structure but resorting to k-connected neighbours, outperforming the approach in \cite{article:CPD_PreservingGlobal_LocalStructures}. 

In \cite{article:Registration_CPD_Bayesian} the authors propose a Bayesian Formulation of CPD (BCPD). Under this setting they guarantee convergence of the algorithm, introduce more interpretable parameters and reduce sensitivity to target rotation. 

Unlike CPD, the transformation is defined as $\mathcal{T}(y_m) = sR(y_m + v_m)+ t$, where $s$ is a scale factor, $R$ a rotation matrix, $t$ a translation vector and $v_m$ the displacement vector representing a non-rigid deformation. They formulate a joint distribution for the target points, but also for explicit correspondence vectors between the two shapes. Besides, the motion coherence is expressed as a prior distribution on the displacement vectors instead of a regularization term and the optimization is not based on EM algorithm but rather on Variational Bayesian inference (VBI). Finally, they provide an  acceleration scheme to reduce computation time of the matrices without loss of registration accuracy.

\subsubsection{Graph Matching}
Correspondence can also be found through graph matching where generally each vertex represents a point in the point cloud. Graph matching methods can be of first, second or higher order. First order methods, using only information about each vertex, have been replaced by higher order methods and are not commonly used at the moment. Second order methods \cite{article:Registration_Graph_Matching_Spectral_relaxation,article:Registration_Graph_Matching_SDP_relaxation}  try to match both vertices and edges, while higher order methods \cite{article:Registration_Graph_Matching_HighOrder_randomwalks,article:Registration_Graph_Matching_HighOrder_tensorblock} include extra information such as angles between vertices and have the advantage of being invariant to scale and affine changes. Both can be formulated as a Quadratic Assignment Problem (QAP), which is NP-hard. Thus, most graph matching approaches are limited to a small number of nodes (at most in the order of hundreds) and are therefore not suitable for this application, since our point clouds have thousands of points.

\subsection{Shape completion}

Given the amount of missing data in our scenario, after the registration step a large percentage of points in the template will have no correspondence in the target. In order to relate the complete shape of the ear with the head, we would like to first complete the missing data. Of course the completion can be helped by some information on the shape but this would require a previous model, taking us back to the chicken-and-egg problem mentioned before. Therefore, we consider three alternatives which entail different levels of prior information on the shape.

\subsubsection{Deformed template from the registration method}

Under this option, we take the transformed template from the registration method and see how well it resembles the original target. The main advantage is that this uses the generic transformation model defined by the registration method and does not require any previous knowledge on the shape. On the other hand, this lack of information is expected to lead to shapes less similar to ears.

\subsubsection{Probabilistic PCA}

An option to counteract this problem is to use a previous model of the ear shape. After we establish correspondence for the existing points, it is possible to predict the remaining ones with Probabilistic PCA as suggested in \cite{article:Probabilistic_PCA}. This of course assumes a previous model for which we will use the PCA model obtained from \cite{article:3DMM_Ears_Dai_Data_augmented_journal_citation}.

After registration we can split the target into $X = (X_a,X_b)$, where $X_b$ are the known points and $X_a$ the missing ones. Our goal is to obtain the full shape $X$ from the known points $X_b$.  We can express the distribution of the shape points as 
\begin{equation*}
\begin{split}
p(X) &= p(X_a,X_b) \\&= \mathcal{N} \Big( \begin{bmatrix}
\mu_a\\\mu_b
\end{bmatrix}  , \begin{bmatrix}
W_a W_a^T & W_a W_b^T \\W_b W_a^T & W_b W_b^T +\sigma I_{3\tilde{N}}
\end{bmatrix}  \Big),
\end{split}
\end{equation*}
where $\mu_a$ and $\mu_b$ are the mean shape points corresponding to $X_a$ and $X_b$, $W_a$ and $W_b$ the matrices with the respective dimensions of the principal components and $I_{3\tilde{N}}$ an identity matrix of size $3\tilde{N}$, where $\tilde{N}$ is the number of known points.

We can then obtain the most likely $\alpha$ with the following expression
\begin{equation*}
p(\alpha\| X_b) = \mathcal{N}(M^{-1} W_b^T\sigma{-2}(X_b-\mu_b), M^{-1}),
\end{equation*}
where $M =\sigma{-2} W_b^T W_b + I_d $. Then, the most likely shape is obtained by taking the mean of the previous distribution and computing $X = W\alpha +\mu$, where $\mu$ and $W$ are respectively the mean shape and principal components of the original model from \cite{article:3DMM_Ears_Dai_Data_augmented_journal_citation}.

\subsubsection{Gaussian Process framework}

The disadvantage of the previous option is that the initial PCA model limits the shape space. It would be ideal to include some shape information but still allow for some freedom in shape matching. A framework contemplating this option is the Gaussian Process presented in \cite{article:Gaussian_Process_2017_Vetter}.

The main idea of this approach is to model a shape $s$ as $s = \{x+u(x) | x \in \Gamma_R \}$, where $\Gamma_R $ is a reference shape and $u$ a deformation acting on the reference shape and resulting in $s$. Then, the deformation $u$ is modelled as a Gaussian Process $u \sim \mathcal{GP} (\mu,k)$, where $\mu $ is the mean deformation and $k$ is a covariance function/kernel. 

A parametric and low dimensional model can be obtained by representing the Gaussian process using the leading functions ($\phi_i$) of its Karhunen-Loève expansion, so that we obtain a model similar to PC
\begin{equation*}
u = \mu + \sum_{i=1}^r \alpha_i \sqrt{\lambda_i} \phi_i,
\end{equation*}
where $\alpha_i \in \mathcal{N}(0,1)$ and $(\lambda_i, \phi_i)$ are the eigenvalue/ eigenfunction pairs of the integral operator $\mathcal{T}_\kappa f(\cdot) \coloneqq \int_\Omega \kappa(x,\cdot) f(x) d \rho(x)$. More details on this low rank approximation model can be found in \cite{article:GP_framework_low_rank_app}.

Therefore, we can include prior shape information when defining the kernel and mean of the Gaussian Process. We can include the dataset information with the mean deformation at each point $x$ given as
\begin{equation*}
\mu_{SSM}(x) = \frac{1}{n}\sum_{i=1}^{n} u_i(x)
\end{equation*}
and the so called sample covariance kernel, where the covariance function $k_{SSM}(x,x')$ at points $x$ and $x'$ is given as
\begin{equation*}
\begin{split}
\frac{1}{n-1}\sum_{i=1}^{n} (u_i(x)-\mu_{SSM}(x) )(u_i(x')-\mu_{SSM}(x') )^T.
\end{split}
\end{equation*}

Then, we can augment this kernel with a Gaussian kernel $g(x,x') = \exp{-\frac{\|x-x'\|^2}{\sigma^2}}$, which merely models smooth deformations. The final kernel is obtained by summing the two previous ones $k_{final}(x,x') =k(x,x')+g(x,x')$, thus increasing the variability of the initial sample kernel.

While this framework can be used on its own to perform registration, it is not particularly suited for the kind of the data we have and if used without any previous steps does not produce adequate results. Therefore, we chose to first perform an initial registration with other methods and then use this framework for shape completion. Essentially, this consists of Gaussian Process regression, where the correspondences found by registration of the previous steps are taken as observations. That is, we compute the observed deformation between matched points and then apply regression to predict the remaining shape.

\section{Results}
\label{sec:results}
\subsection{Initial registration}
\label{sec:init_reg}

\begin{figure*}[ht]
	
	\subfloat[]{
		\includegraphics[width=.5\textwidth,clip, trim={80pt 405pt 100pt 230pt}]{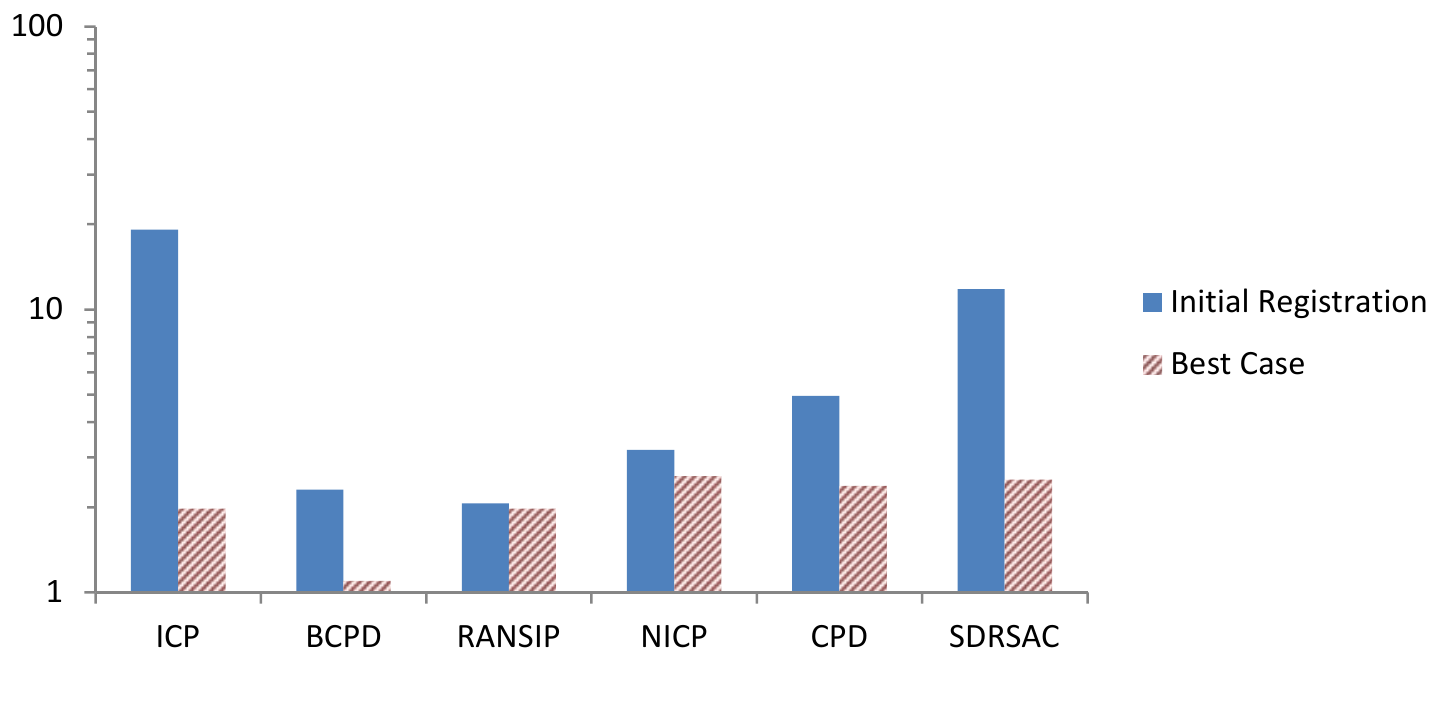}}
	\label{fig:init_reg_a}
	~
	\subfloat[]{
		\includegraphics[width=.5\textwidth,clip, trim={80pt 280pt 50pt 300pt}]{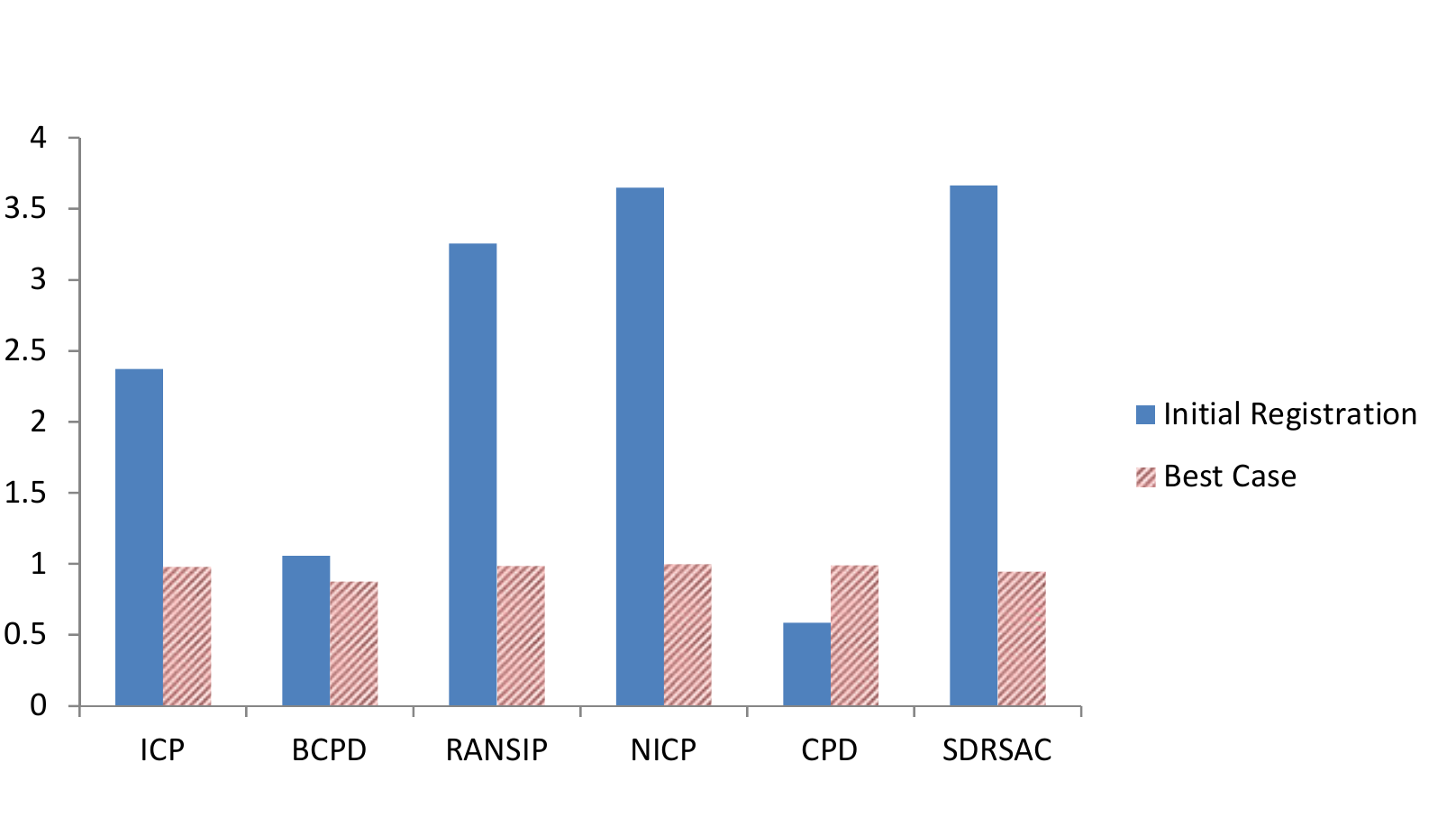}}
	\caption{Distance error (a) and fraction of correspondences (b) for the initial registration with each of the different methods. Comparison between the best case scenario and the real data.}
	\label{fig:init_reg_dist}
\end{figure*}

\begin{figure*}[htp]
	\subfloat[]{
		\includegraphics[width=.5\textwidth,clip, trim={50pt 455pt 150pt 150pt}]{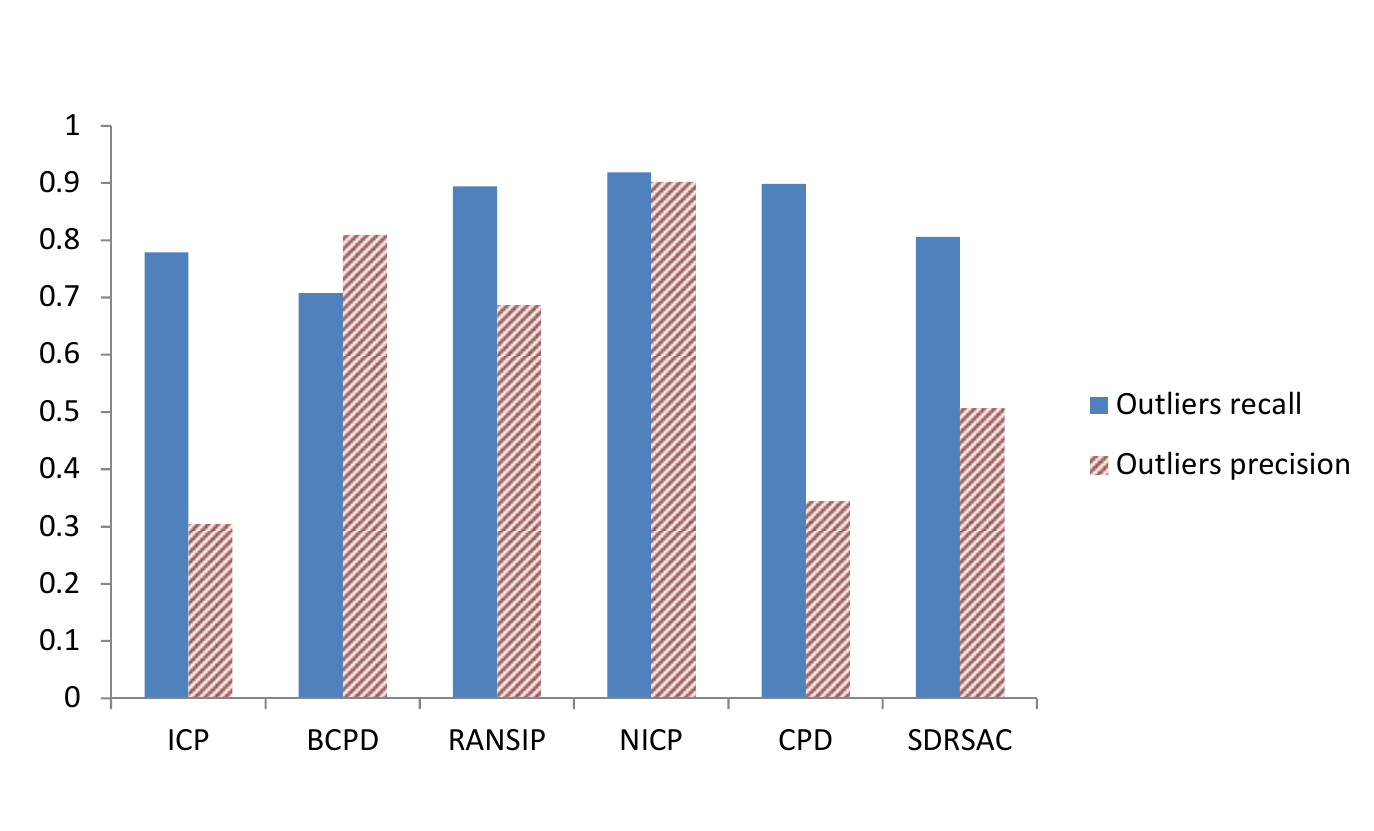}}
	~
	\subfloat[]{
		\includegraphics[width=.5\textwidth,clip, trim={50pt 460pt 150pt 150pt}]{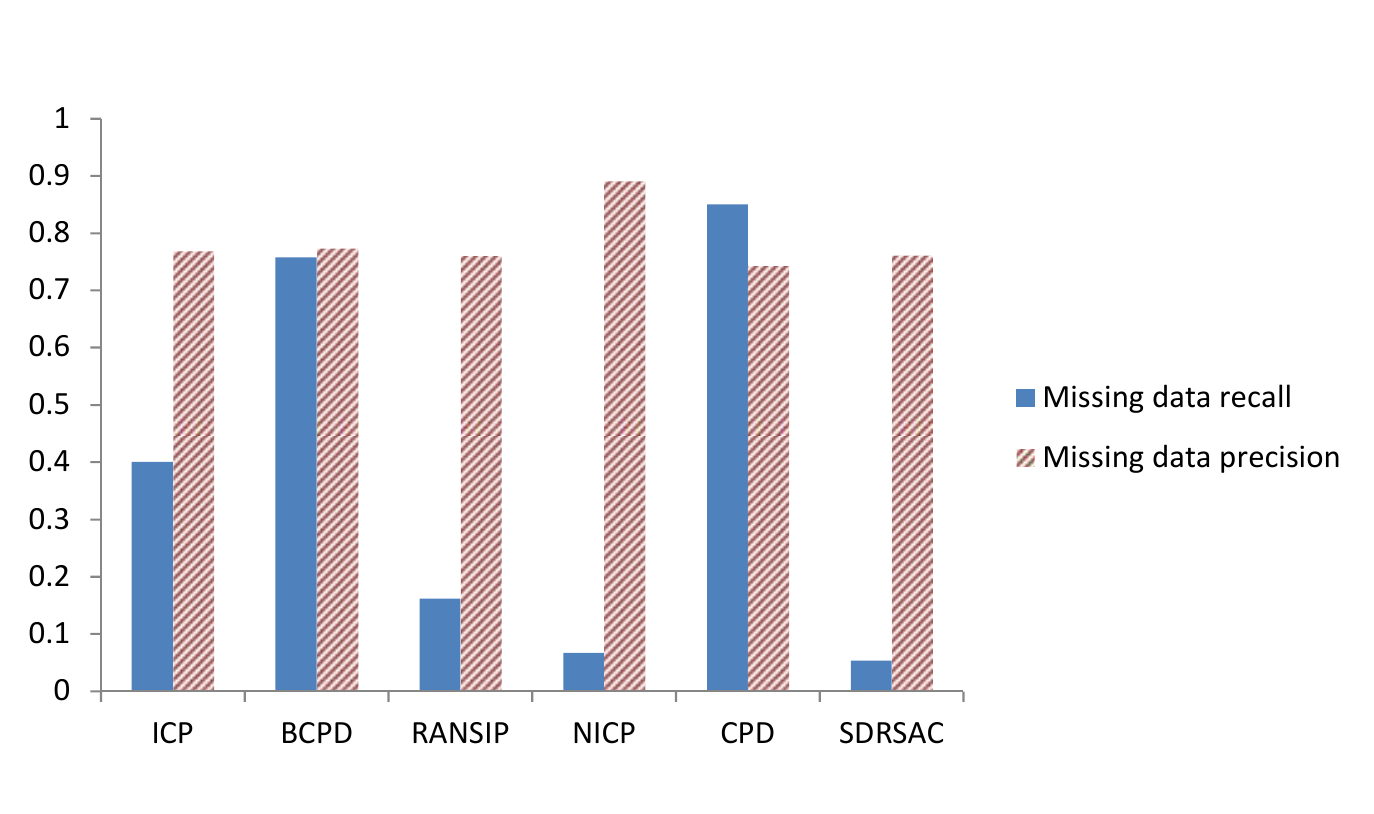}}
	
	\caption{Outlier (a) and missing data (b) metrics for the initial registration with  each of the different methods. The results are only with respect to the real case scenario, since the best case does not have any outliers or missing data. }
	\label{fig:init_reg_out_miss}
\end{figure*}

In this section, we apply different state-of-the-art registration methods to the Simulated dataset. The goal is to roughly register the samples, ideally keeping all the points belonging to the ear and identifying the outliers. We choose methods from different areas, both rigid and non-rigid, to understand which ones are more suited to overcome the data problems. Two fundamental methods are ICP \cite{article:Registration_ICP} for the rigid registration and CPD \cite{article:Registration_CPD} for the non-rigid counterpart. These methods are considered state-of-art but not particularly suited to the data problems we find in this situation. Therefore, we chose more recent variations for both cases: NICP\cite{article:Registration_NICP_2007} (the non-rigid version of ICP) and BCPD \cite{article:Registration_CPD_Bayesian} (the Bayesian formulation of CPD). Given the susceptibility of ICP-like method to local optima, we also test two RANSAC methods: the SDRSAC \cite{article:SDRSAC} and our proposed approach RANSIP.

For each of the methods and scenarios an extensive study was performed in order to find the optimal parameters. The results here presented correspond to the best performance for each step.

Furthermore, we compare the results with a so called \textit{best case scenario}. This consists of registering the original Ear dataset without missing data, outliers or noise, with the template in the correct position. For this case, we know that there should be a one-to-one correspondence for every point.

Fig.~\ref{fig:init_reg_dist} provides the general metrics for this step, comparing the several methods both for the real and best case scenario. Fig.~\ref{fig:init_reg_out_miss} depicts the outlier and missing data metrics only for the real case, as they are not applicable in the other.

\subsubsection{ICP}

It is immediately evident that ICP does not cope well with the real case scenario. This is natural, since ICP can easily be trapped in local optima and the prevalence of data problems creates more local optima. That is, with ICP the template easily falls into positions far away from the ground truth.

However, for the best case scenario, where template is in the correct initial position and there is a one-to-one correspondence we expect ICP to perform well, and that is indeed the case. In fact, the distance error for this case (around $2mm$) is an important value as it tells us that limited to rigid deformations this is the best error we can expect to find and it is due to the shape differences between the template and targets.

Regarding the correspondences fraction, it is noted that the original version of ICP (used here) allows for a point of the target to be associated with more than one point of the template. Consequently, in the best case, where both point sets have the same number of points, the fraction is around 1. In the real case, if a point in the template does not have a correspondence on the target (missing data), it will be associated to another point as long as is stays within the defined threshold, even if the target point was already associated to another template point.  This means we will find more correspondences than we should and explains the value over 1.

The latter conclusion is further proved by the missing data recall of ICP in Fig.~\ref{fig:init_reg_out_miss}. The low value for this metrics expresses the restricted ability in detecting missing points.

\subsubsection{RANSAC methods}

Both SDRSAC and RANSIP are composed of several runs of ICP. Therefore, without any data problems they should have a similar performance as the latter, as indeed is evidenced by Fig.~\ref{fig:init_reg_dist} for the modified version. The original one has a higher distance error and this could be explained by the fact that we are introducing a random initialization when the original one was already good. Meaning that the cost considered by the SDRSAC is not the most adequate for our data, as it does not detect the best transformation as such.

Looking at the real case, the fact that ICP is run several times helps in avoiding the local optima trapping, leading to an increased performance for the initial registration, as shown by the distance error of RANSIP. The correspondence fraction is still considerably above 1, for the reasons already stated with respect to ICP. When it comes to SDRSAC the performance is still better than the ICP, but again it is evident that the cost is not the most adequate for this situation.

Regarding Fig.~\ref{fig:init_reg_out_miss}, we notice that both outlier metrics and missing data precision are close to 1 (being this the desired value), while missing data recall is very low. This is again related to the correspondence fraction as explained for ICP. However, since the goal of this first step is to remove the majority of outliers, this low value does not prevent the use of the method.

\subsubsection{CPD}

For the best case scenario, CPD performs slightly worse than rigid methods such as ICP or SDRSAC. In this case, the non-rigid deformation does not seem to help in the registration, which could be due to the fact that the dataset does not have enough variability. That is, as the data was sampled from the model, the shapes are considerably similar amongst themselves and rigid deformations to the template are enough for a small distance error. 

CPD does not handle well the real data. Despite the low value of distance error when compared with ICP, we see that this is achieved due to the low fraction of correspondences found (Fig.~\ref{fig:init_reg_dist}). This is in accordance with the low outlier precision in Fig.~\ref{fig:init_reg_out_miss}, showing that a high number of non-outliers is being identified as outliers.

\subsubsection{BCPD}

Unlike CPD, BCPD seems to improve the distance error when compared to the non-rigid methods, which would disprove the justification given for the CPD increase. However, the correspondence fraction for BCPD is lower than for the others (around $90\%$), which means it is wrongly discarding points as outliers, thus obtaining a lower distance error. 

When it comes to the real case, BCPD performs well, achieving a distance error slightly above the best case of the rigid methods and even below the best case of CPD and NICP. This does not come at the cost of the correspondence fraction as this value stays close to 1 (Fig.~\ref{fig:reg_ref_corr}).

\subsection{Registration refinement}
\label{sec:reg_refinement}

The output from the first step allows us to remove the majority of the outliers and only keep the non-outliers identified by each method. Given the previous results, we take only the BCPD and RANSIP outputs. With this clean data it is expected than most of the methods have a better performance and so we test each of the previous approaches for both BCPD and RANSIP outputs.

\begin{figure}[h]
	\includegraphics[width=.5\textwidth,clip, trim={100pt 460pt 120pt 150pt}]{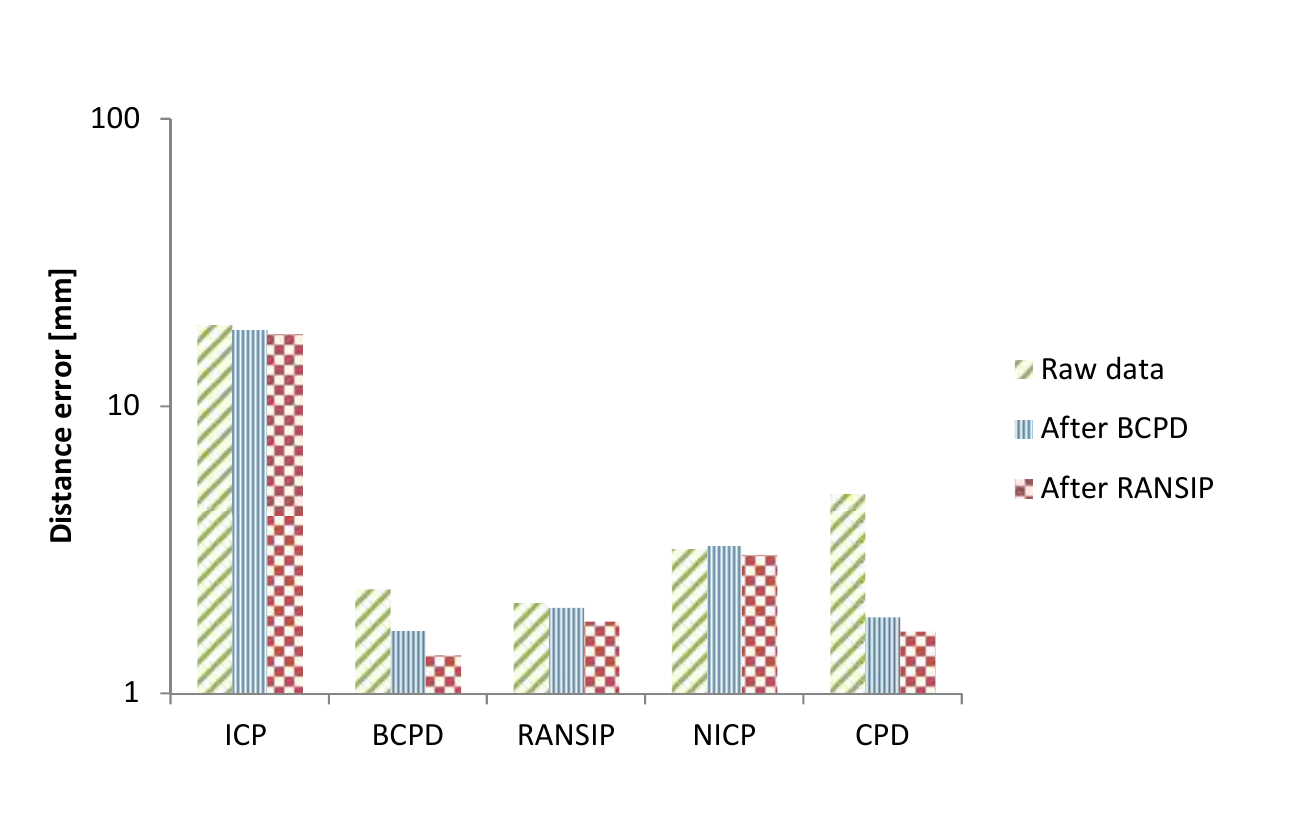}	
	\caption{Distance error for the registration refinement with each of the different methods. Comparison between registration of the initial data, the data after outlier removal with BCPD and with RANSIP. }
	\label{fig:reg_ref_dist}
\end{figure}

\begin{figure}[ht]
	\includegraphics[width=.5\textwidth,clip, trim={60pt 520pt 60pt 150pt}]{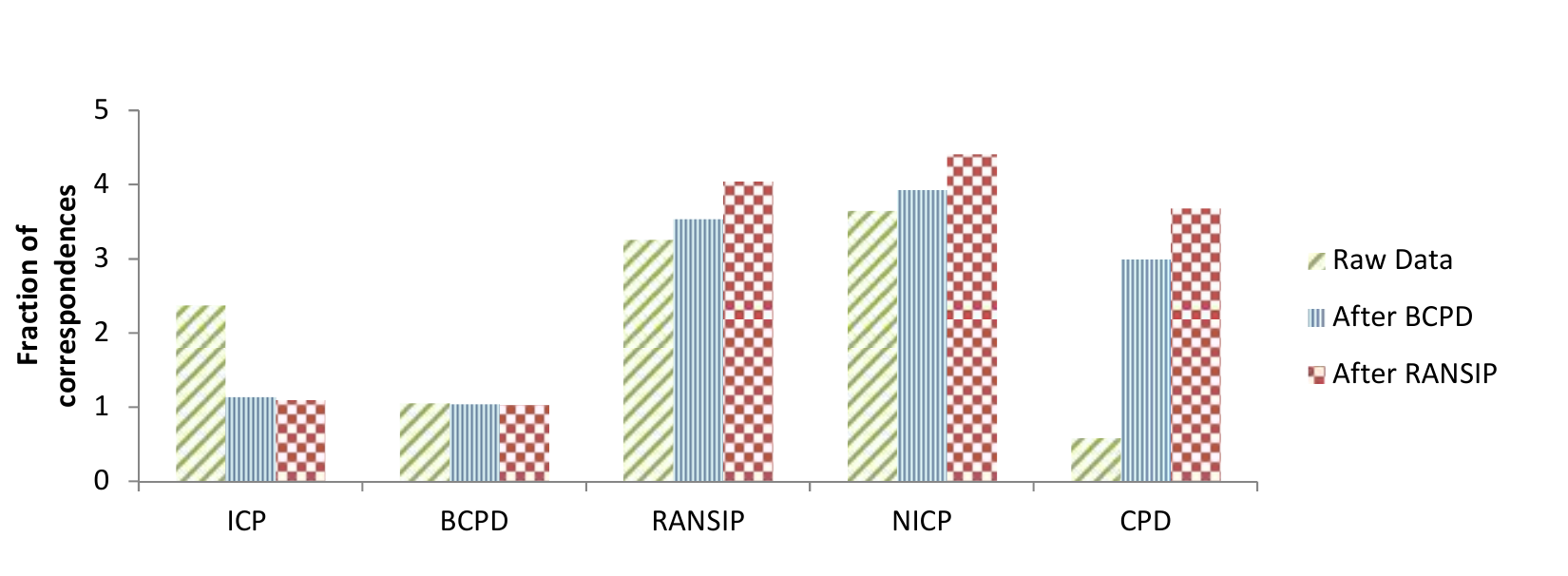}
	\caption{ Fraction of correspondences for the registration refinement with  each of the different methods. Comparison between registration of the initial data, the data after outlier removal with BCPD and with RANSIP. }
	\label{fig:reg_ref_corr}
\end{figure}

\begin{figure*}[ht]
	\subfloat[]{
		\includegraphics[width=.5\textwidth,clip, trim={50pt 520pt 60pt 80pt}]{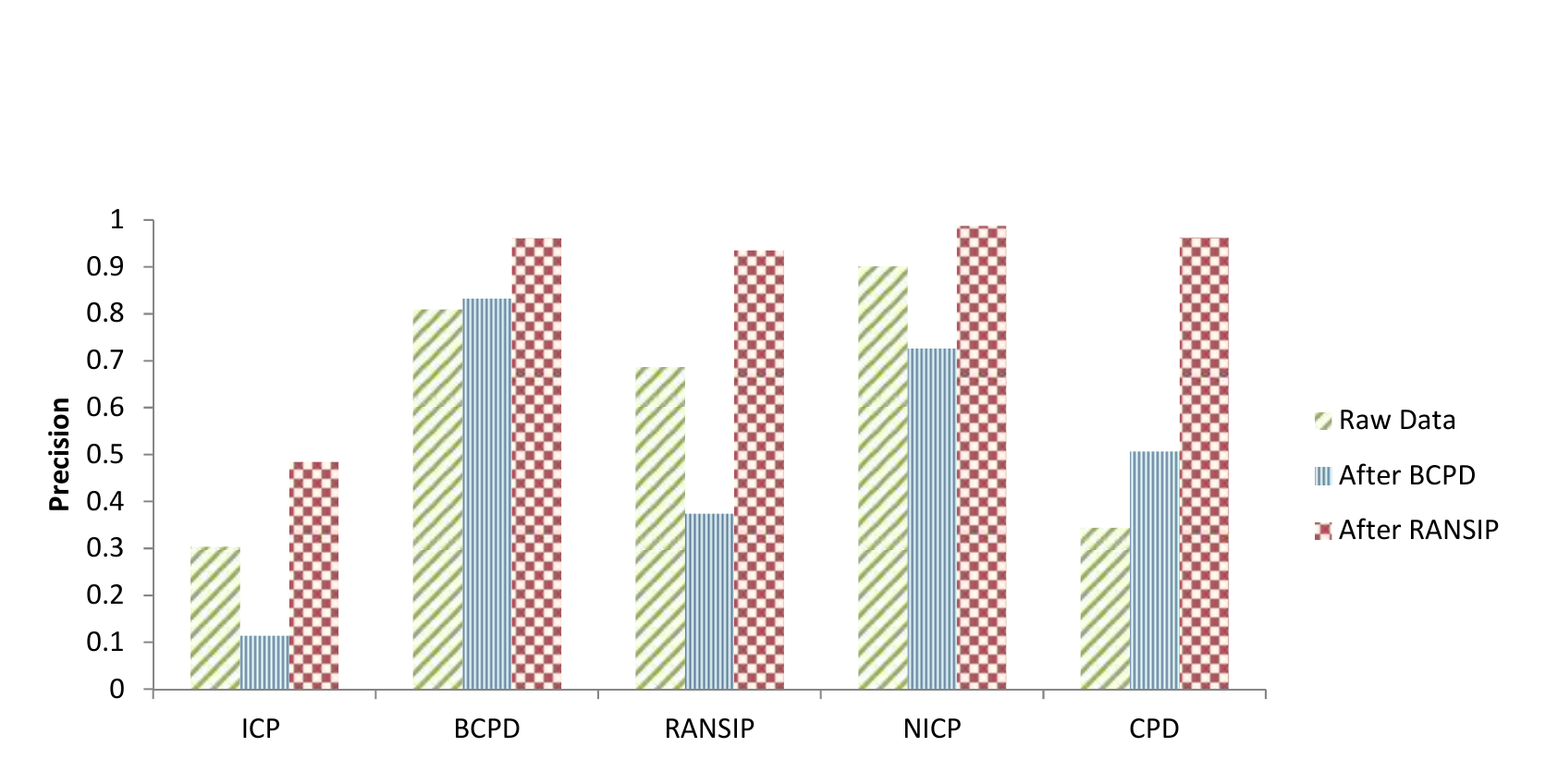}}
	~
	\subfloat[]{
		\includegraphics[width=.5\textwidth,clip, trim={50pt 300pt 40pt 300pt}]{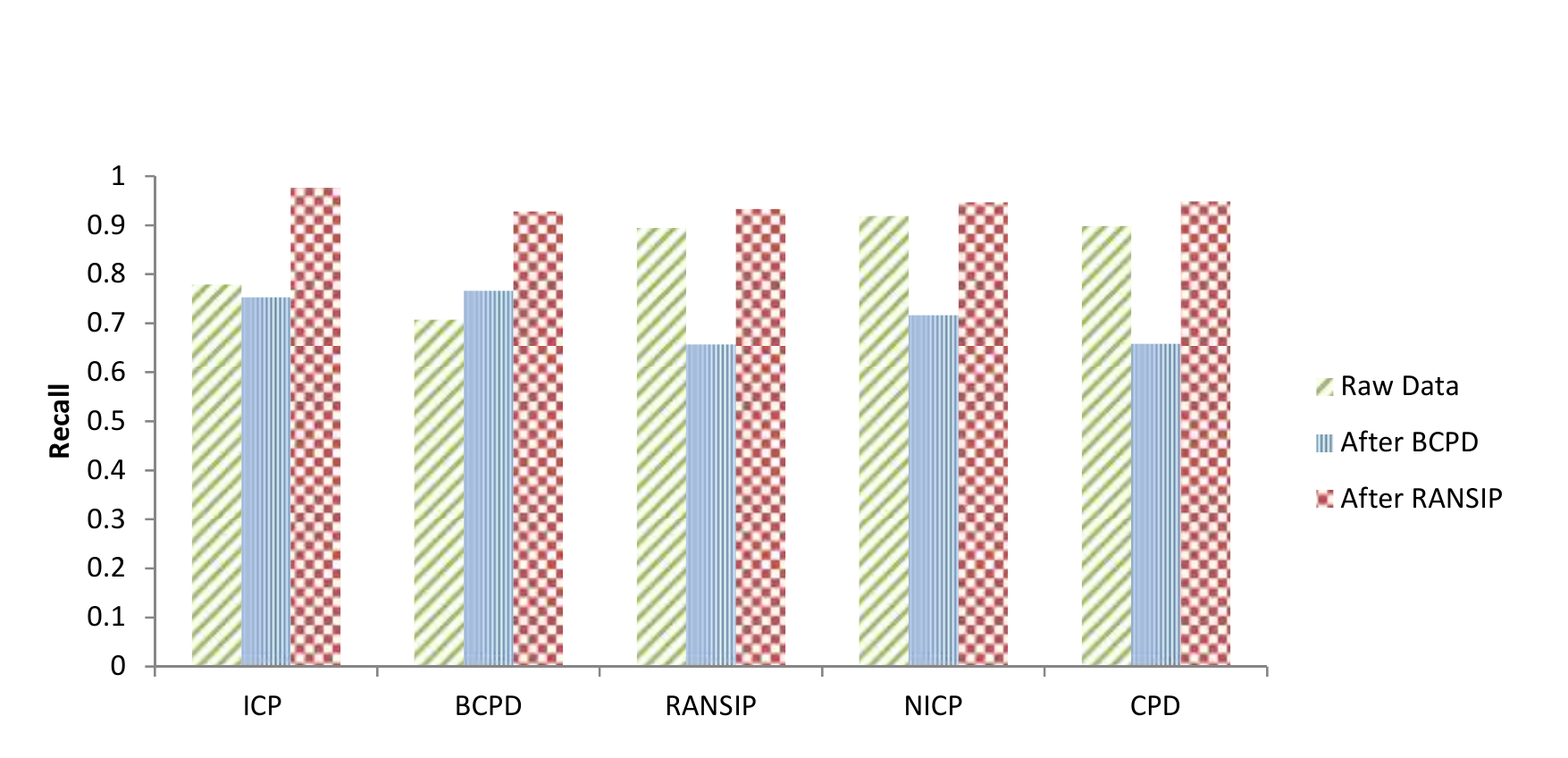}}	
	\caption{Outlier metrics (precision on the left and recall on the right), for the registration refinement with  each of the different methods. Comparison between registration of the initial data, the data after outlier removal with BCPD and with RANSIP.}
	\label{fig:ref_reg_out}
\end{figure*}

\begin{figure*}[htp]
	\subfloat[]{
		\includegraphics[width=.5\textwidth,clip, trim={50pt 300pt 50pt 320pt}]{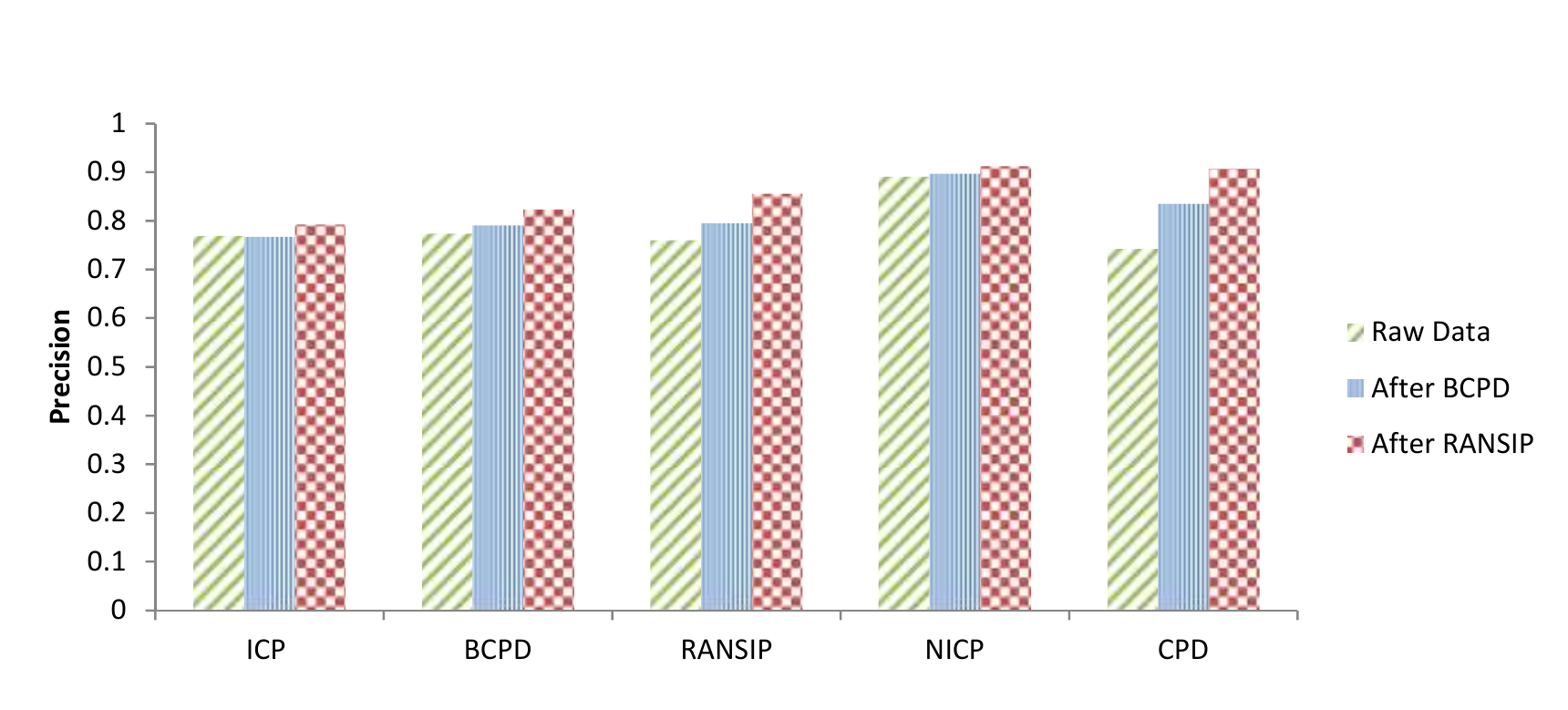}}
	~
	\subfloat[]{
		\includegraphics[width=.5\textwidth,clip, trim={50pt 300pt 50pt 320pt}]{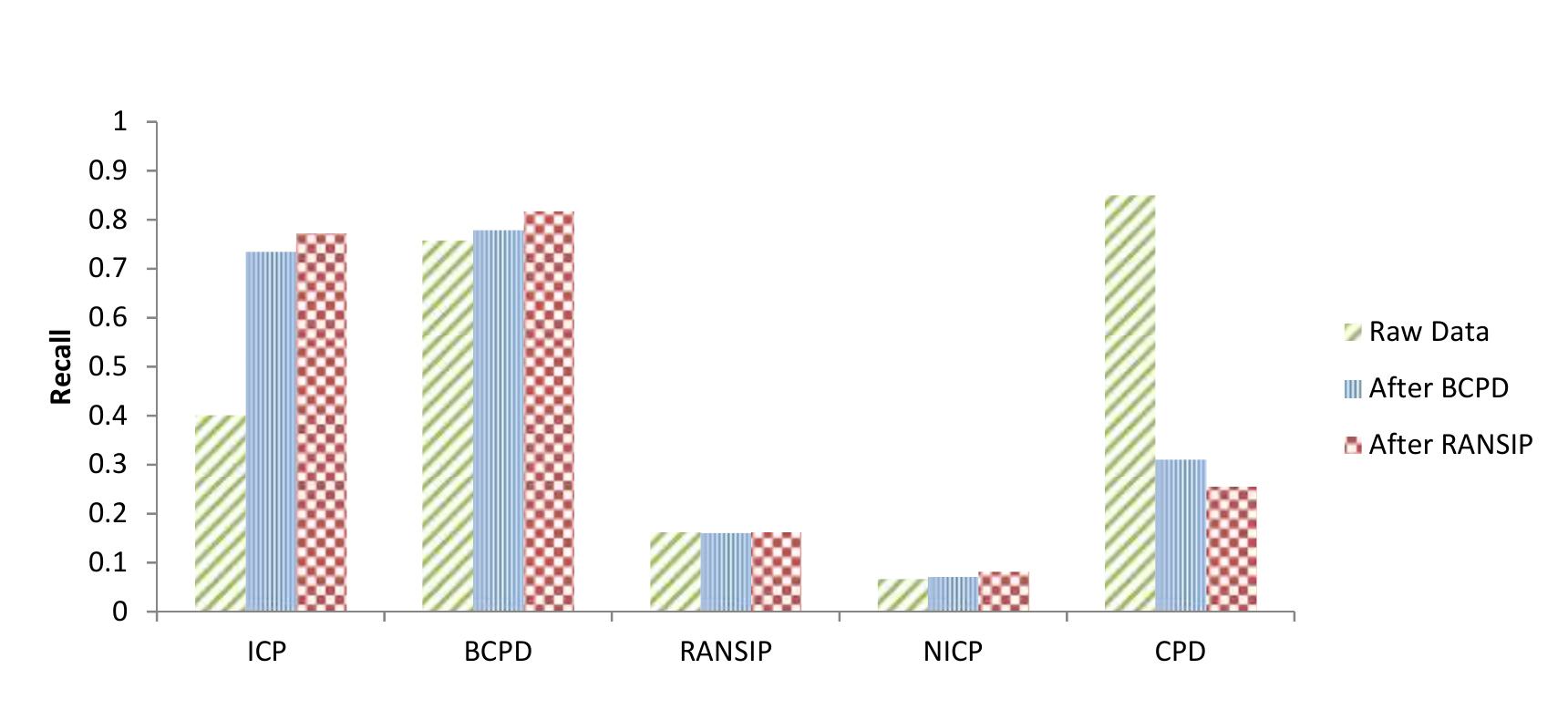}}
	
	\caption{Missing data metrics (precision on the left and recall on the right), for the registration refinement with  each of the different methods.  Comparison between registration of the initial data, the data after outlier removal with BCPD and with RANSIP. }
	\label{fig:ref_reg_miss}
\end{figure*}

Fig.~\ref{fig:reg_ref_dist} shows the distance error, Fig.~\ref{fig:reg_ref_corr} the fraction of correspondences, Fig.~\ref{fig:ref_reg_out} and Fig.~\ref{fig:ref_reg_miss} the outlier and missing data metrics, respectively.

Regarding ICP, we note that even with the removal of the outliers, the remaining data problems still prevent the method from achieving an acceptable performance. CPD, on the other hand, benefits from this removal and is able to express lower values for the distance error, although it has a correspondence fraction considerably above 1. 

For almost any metric and any subsequent method, we notice that RANSIP in the first step produces the best performance. Consequently, this is selected as the method for the first step of the pipeline.

Regarding the choice for the second step, we are particularly interested in achieving good performance on the missing data metrics, as it will allow us to correctly identify where to do shape completion in the subsequent step. Looking at Fig.~\ref{fig:ref_reg_miss}, we conclude that the precision is similar across the different strategies, meaning that non-missing points are correctly identified as such, so we are not loosing important information. However, when it comes to the ability to detect the true missing points, expressed by the recall, most approaches present low values, with the exception of ICP and BCPD. 

The comparison between CPD and BCP is pertinent. After the outlier removal, CPD manages to achieve competitive values of distance error with respect to BCPD. However, it is not able to cope well with missing data and ends up attributing correspondences to template points that should not have one. This increases the fraction of correspondences and decreases the recall for missing data.

\subsection{Shape completion}
\label{sec:reg_completion}

\begin{figure}[h]
	\includegraphics[width=.5\textwidth,clip, trim={60pt 450pt 100pt 180pt}]{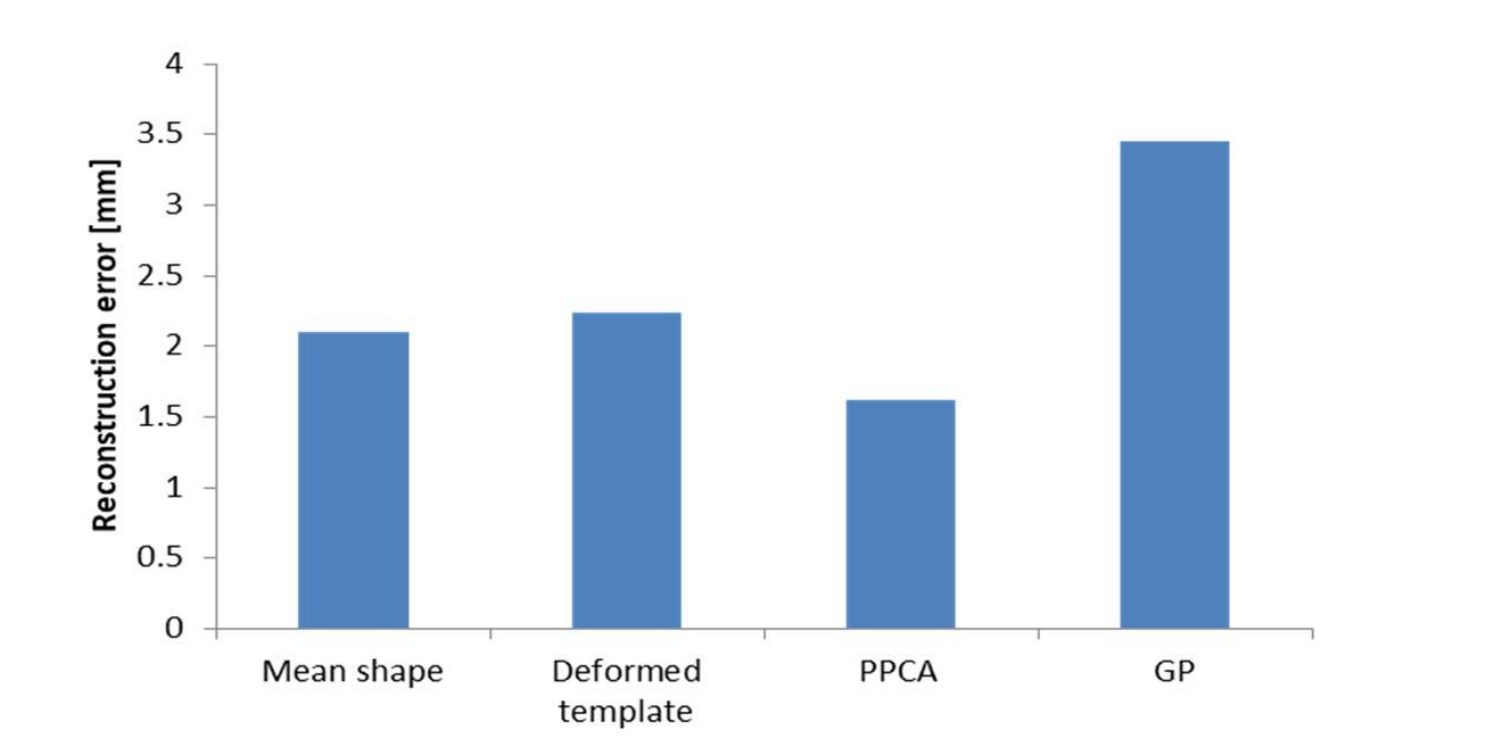}
	\caption{ Comparison of the reconstruction error with the different options. The error with respect to the mean shape is used as baseline.}
	\label{fig:shape_completion}
\end{figure}

For this step we take approximately $10\%$ of the dataset as shapes to be predicted and use the remaining shapes to build the models. Fig.~\ref{fig:shape_completion} shows the reconstruction error for each of the three alternatives described. We consider as baseline the mean shape used for the reconstruction of each shape. We can see that the only option with lower error with respect to the baseline is the PPCA. The fact is that given the way the dataset was built, the samples do not present an increased variability. This also explains why the GP approach performs so poorly, since we increased the variability of the model with the Gaussian kernel but we are predicting shapes which are very similar to our PCA model. However, this does not mean that in the real case the GP will perform worse, as we expect the ear from the Head dataset to present more variability. For this reason, in the next section we apply the whole pipeline to the latter dataset.

\subsection{Real data}

\begin{figure*}[htp]
	\subfloat[Deformed template from registration]{
		\includegraphics[width=.3\textwidth,clip, trim={80pt 330pt 120pt 100pt}]{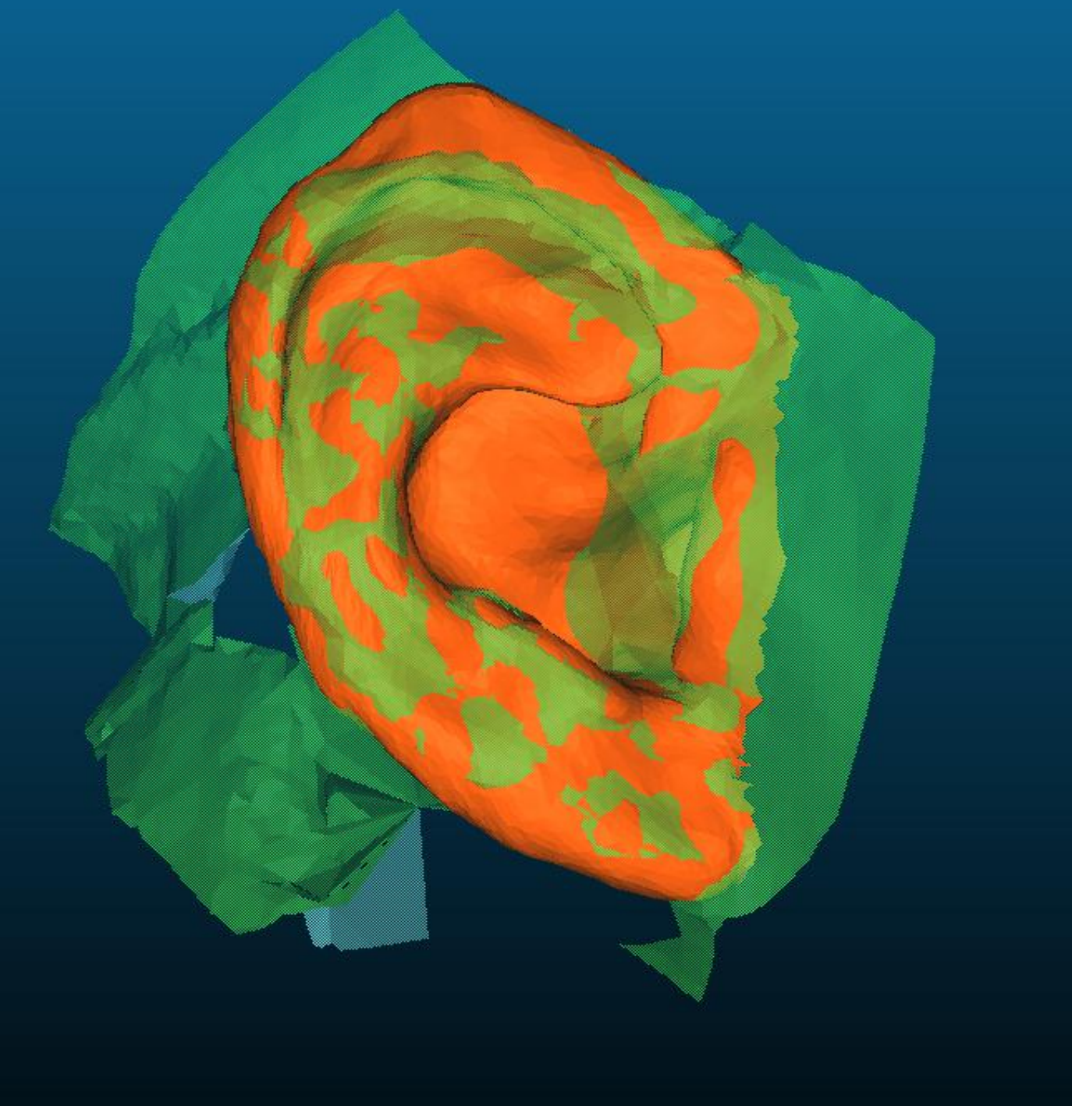} }
	~
	\subfloat[PPCA]{
		\includegraphics[width=.3\textwidth,clip, trim={80pt 350pt 140pt 100pt}]{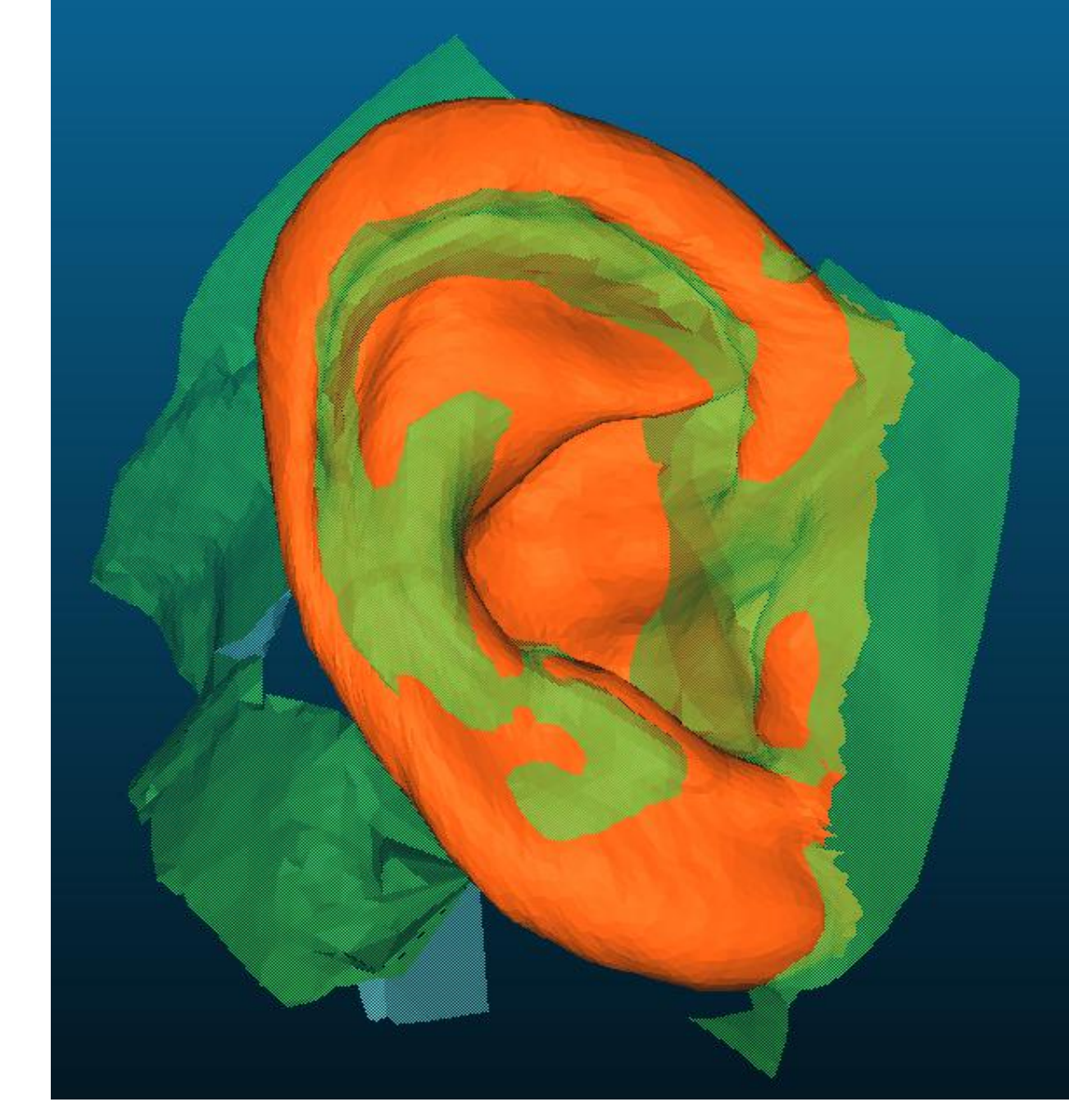}}
	~
	\subfloat[GP]{
		\includegraphics[width=.3\textwidth,clip, trim={100pt 330pt 100pt 100pt}]{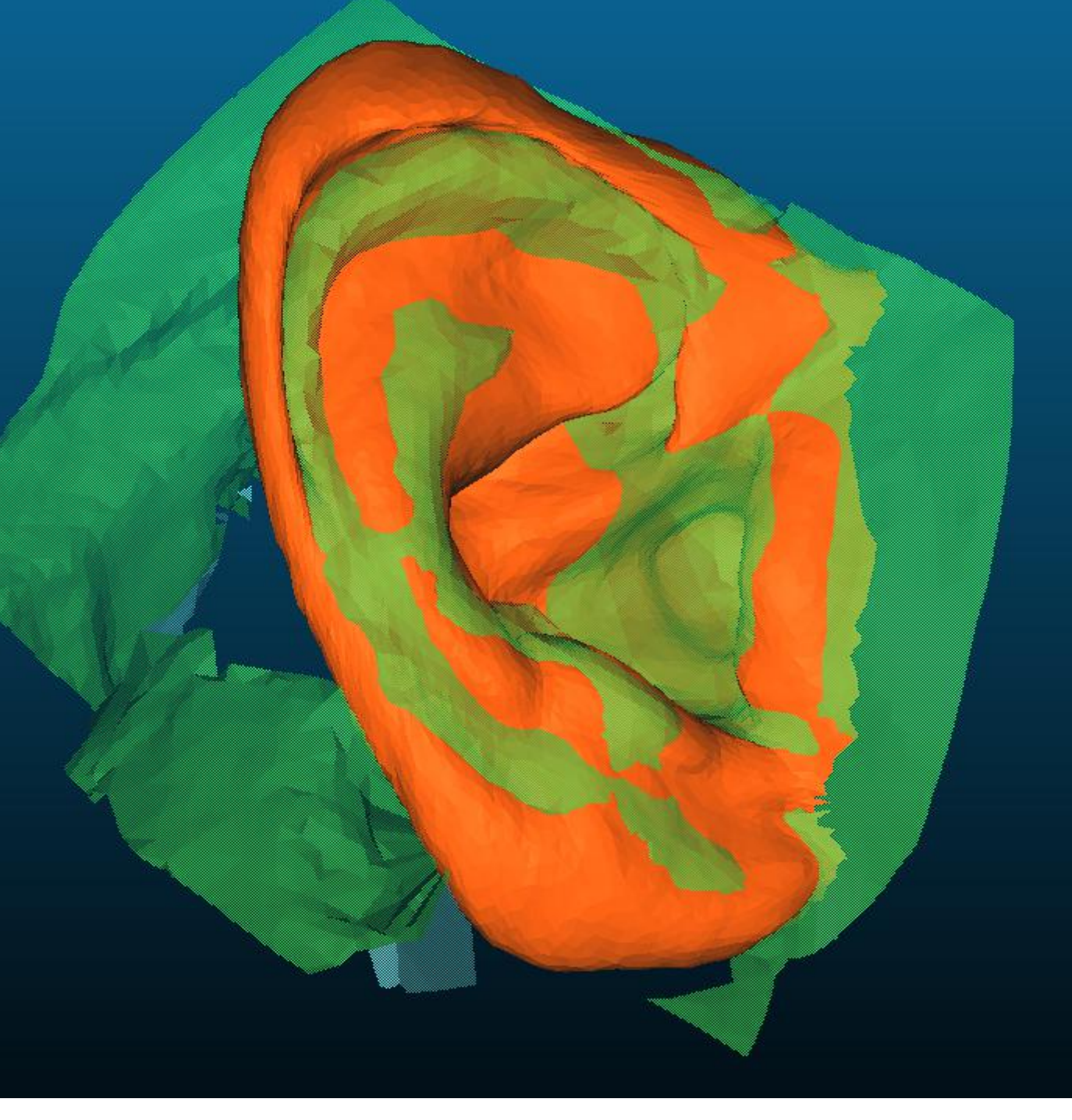}}
	\newline
	
	\subfloat[Deformed template from registration]{
		\includegraphics[width=.3\textwidth,clip, trim={80pt 320pt 100pt 100pt}]{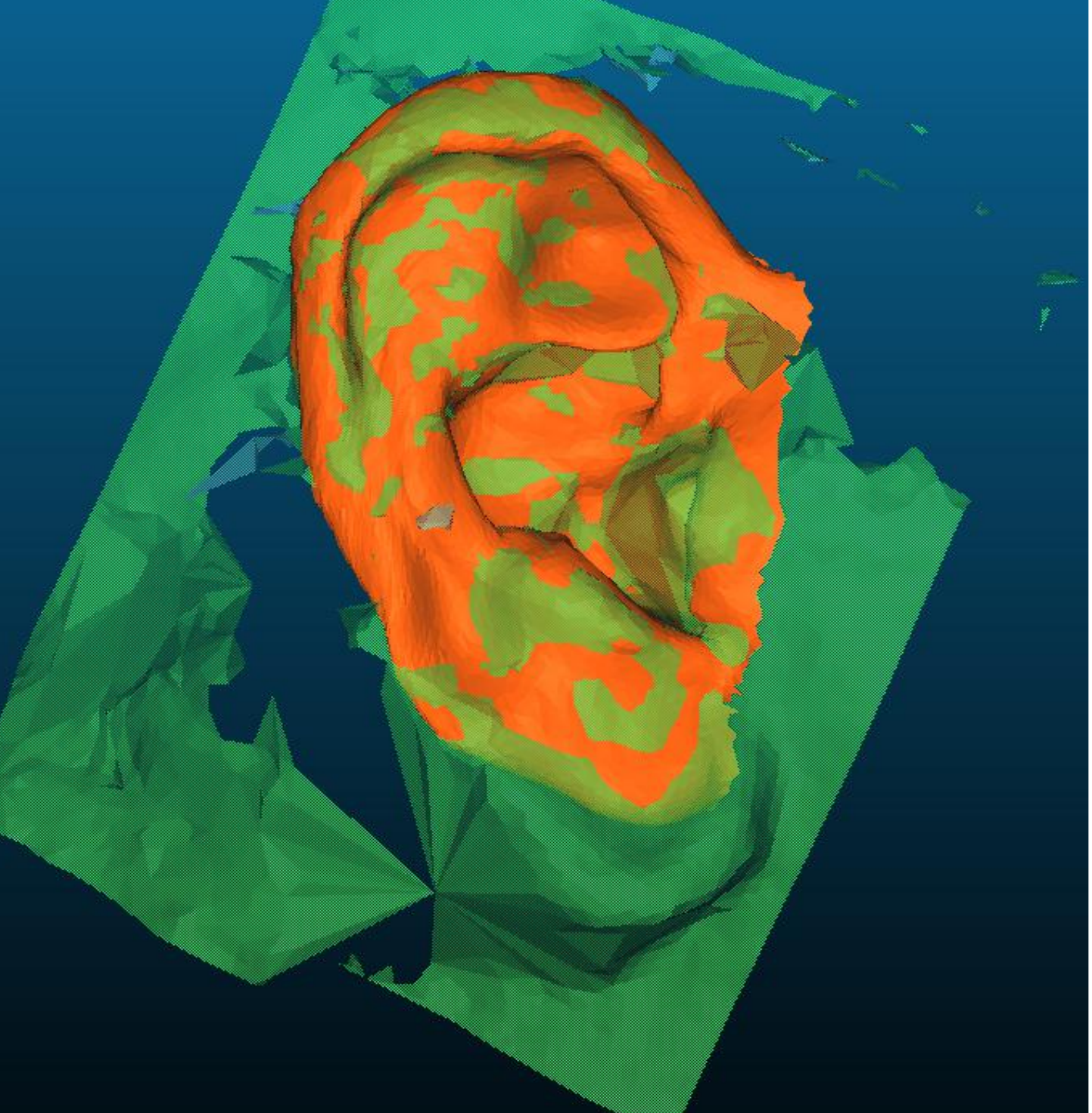}}
	~
	\subfloat[PPCA]{
		\includegraphics[width=.3\textwidth,clip, trim={80pt 340pt 180pt 160pt}]{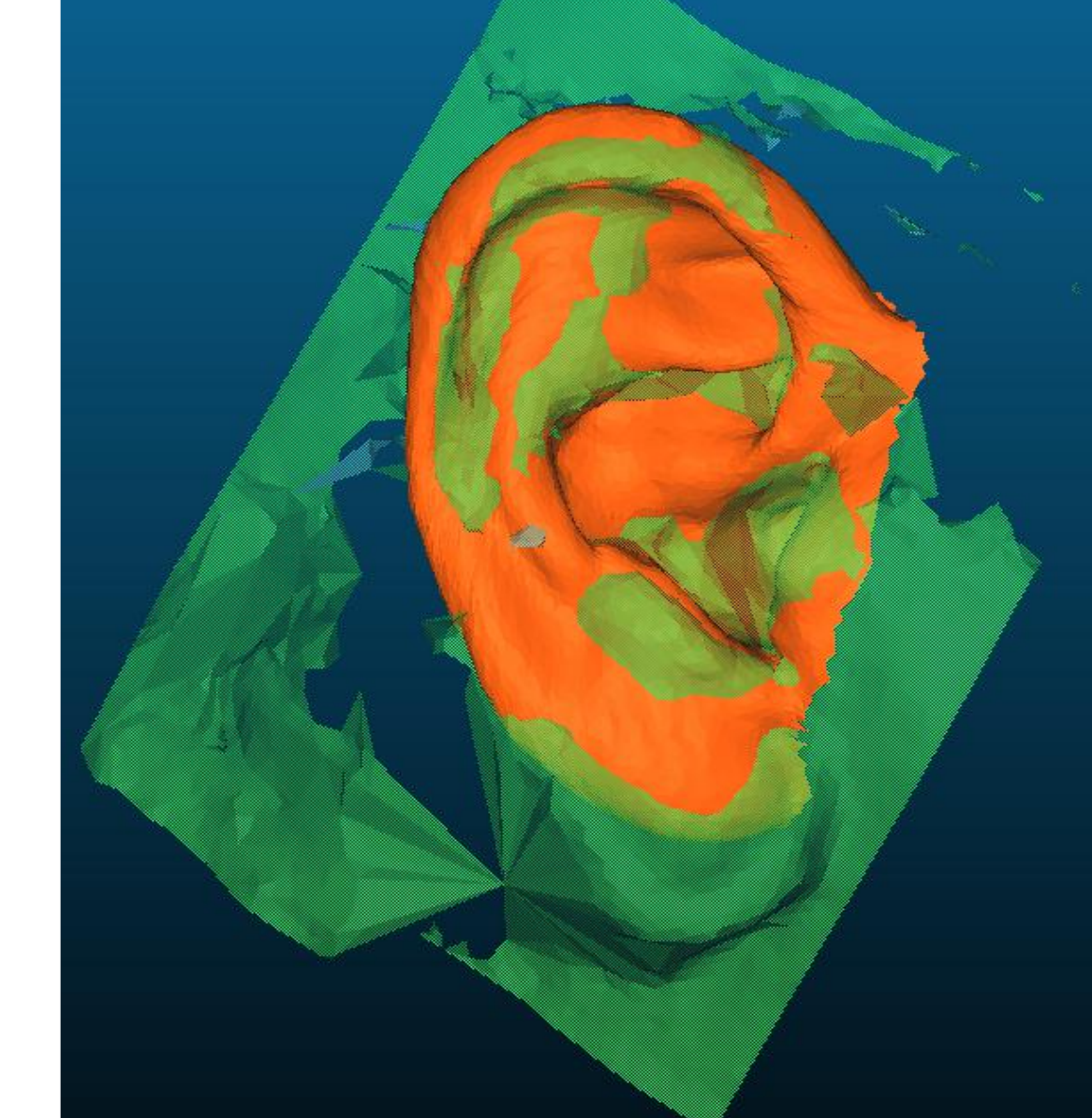}}
	~
	\subfloat[GP]{
		\includegraphics[width=.3\textwidth,clip, trim={80pt 360pt 140pt 100pt}]{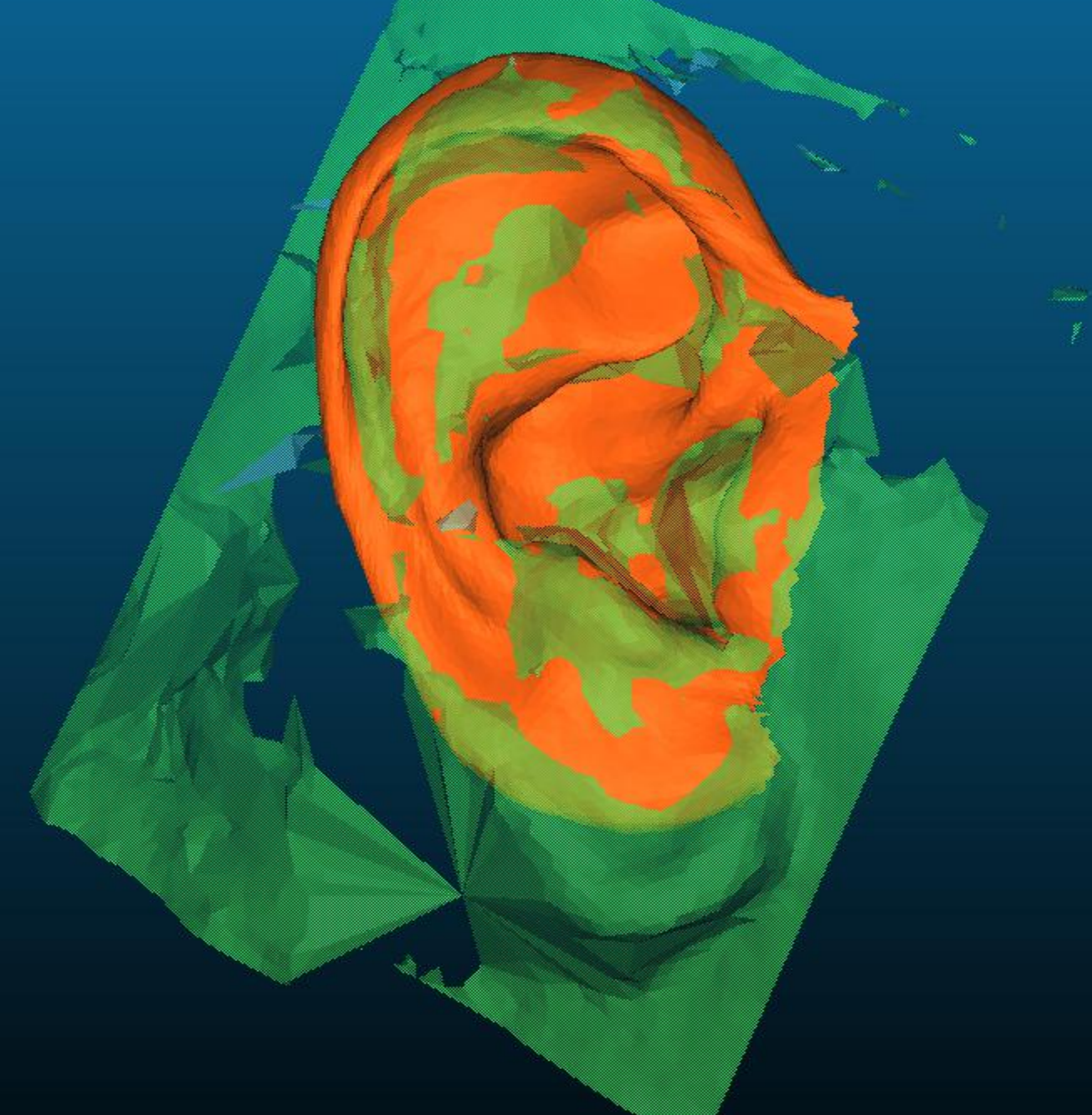}}	
	\newline
	\subfloat[Deformed template from registration]{
		\includegraphics[width=.3\textwidth,clip, trim={100pt 330pt 120pt 140pt}]{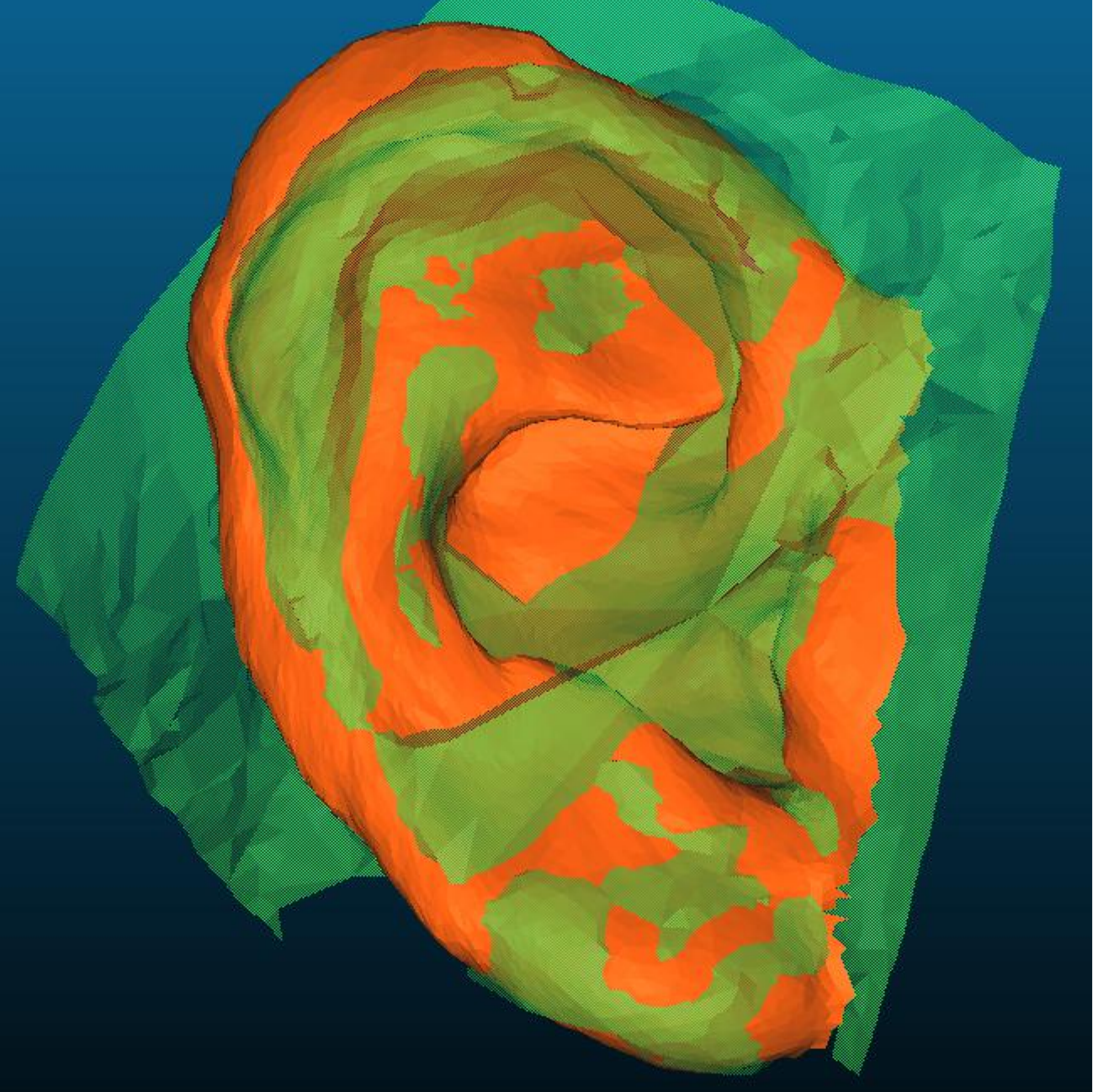}}
	~
	\subfloat[PPCA]{
		\includegraphics[width=.3\textwidth,clip, trim={60pt 250pt 140pt 200pt}]{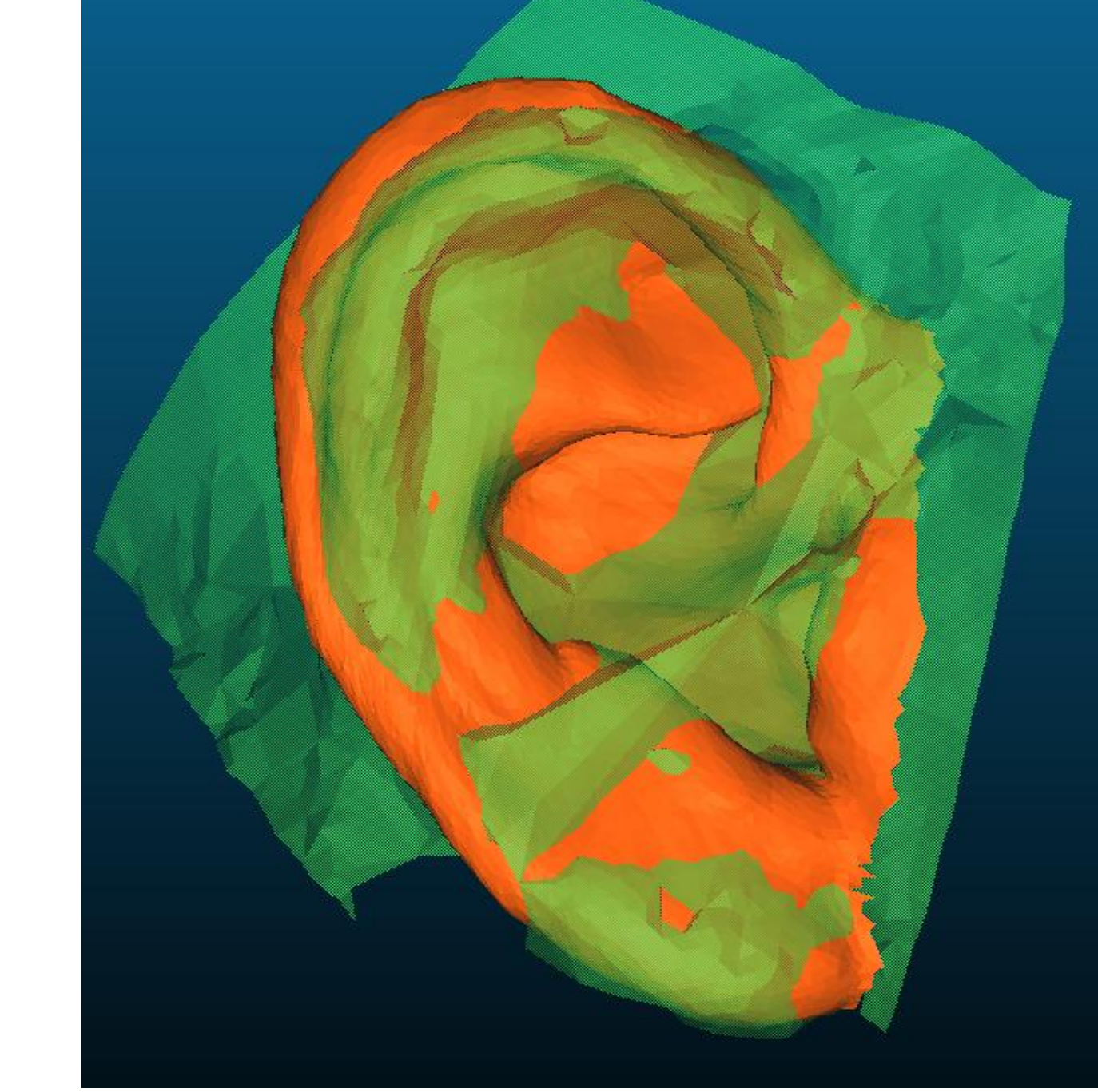}}
	~
	\subfloat[GP]{
		\includegraphics[width=.3\textwidth,clip, trim={130pt 340pt 100pt 140pt}]{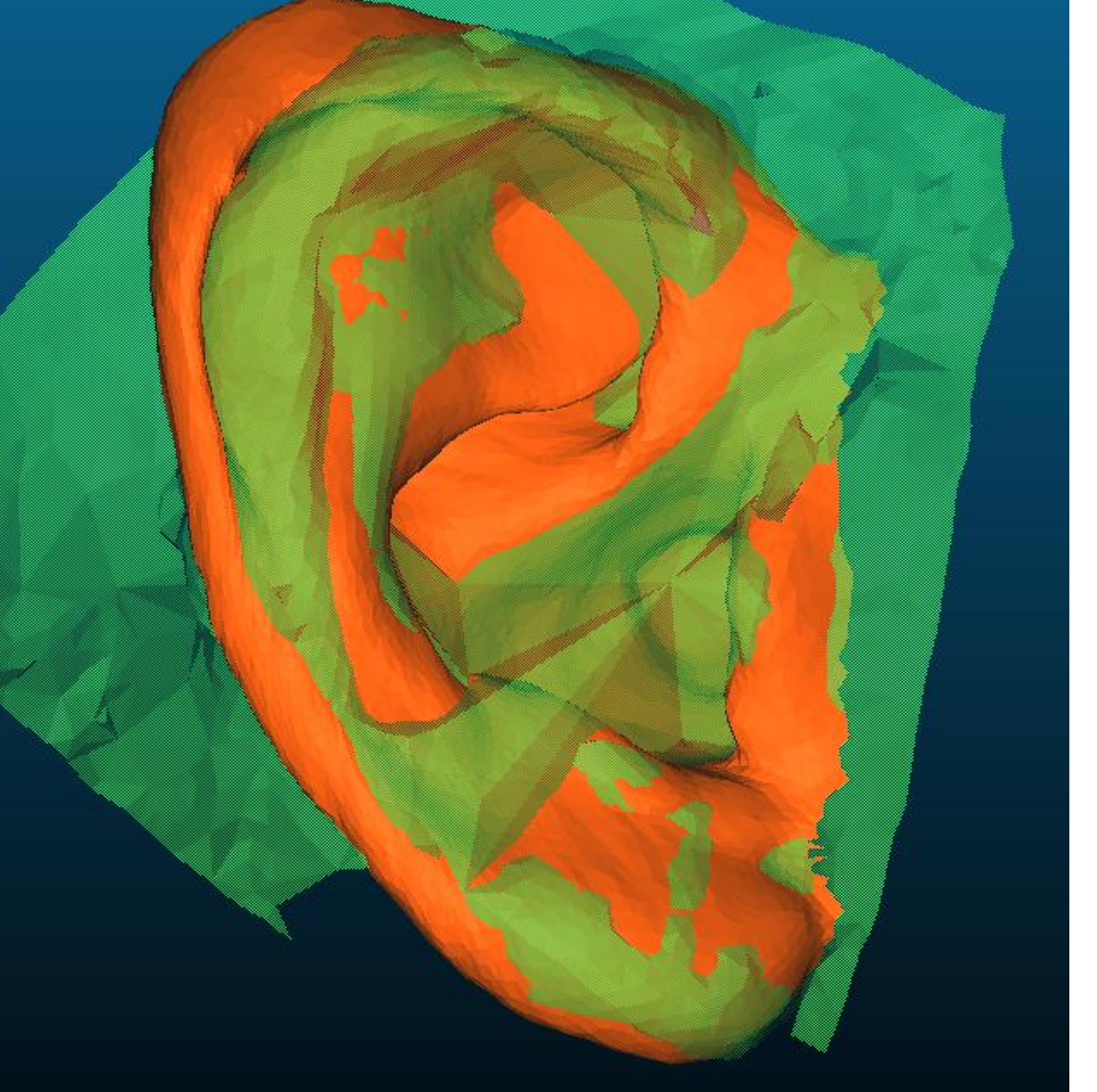}}
	\caption{Example of application of the full pipeline to three real scans of the ear. Each rows depicts the results of a different scan with the shape completion by deformed template from the registration (left column), PPCA (middle column) and GP regression (right column). In each picture, the green shape corresponds to the original scan and the orange one to the shape predicted by the respective approach.}
	\label{fig:real_pipeline}
\end{figure*}

Finally, we apply the complete pipeline to real ears from the Head dataset. Evidently there is no ground truth, so it is not possible to obtain quantitative metrics. Therefore, we evaluate the results by empirical observation in Fig.~\ref{fig:real_pipeline}.

\begin{figure*}[htp]
	\subfloat[]{
		\includegraphics[width=.5\textwidth,clip, trim={60pt 530pt 60pt 140pt}]{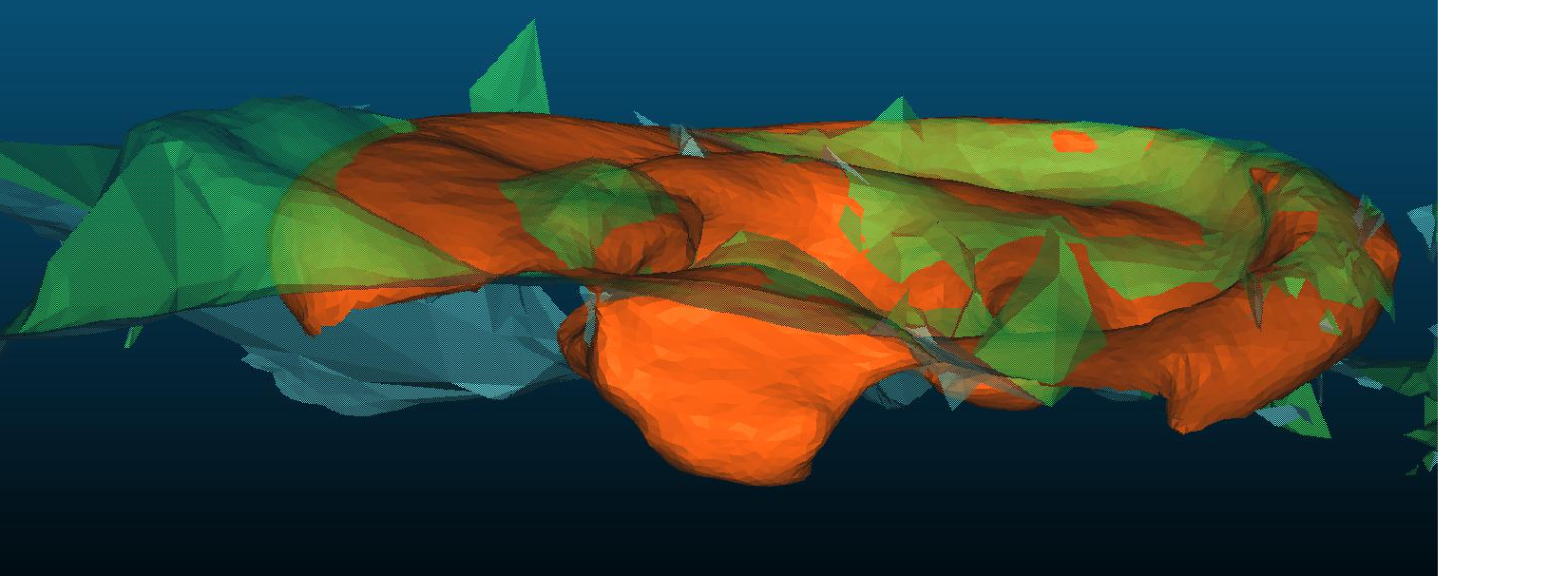}}
	~
	\subfloat[]{
		\includegraphics[width=.5\textwidth,clip, trim={80pt 550pt 70pt 130pt}]{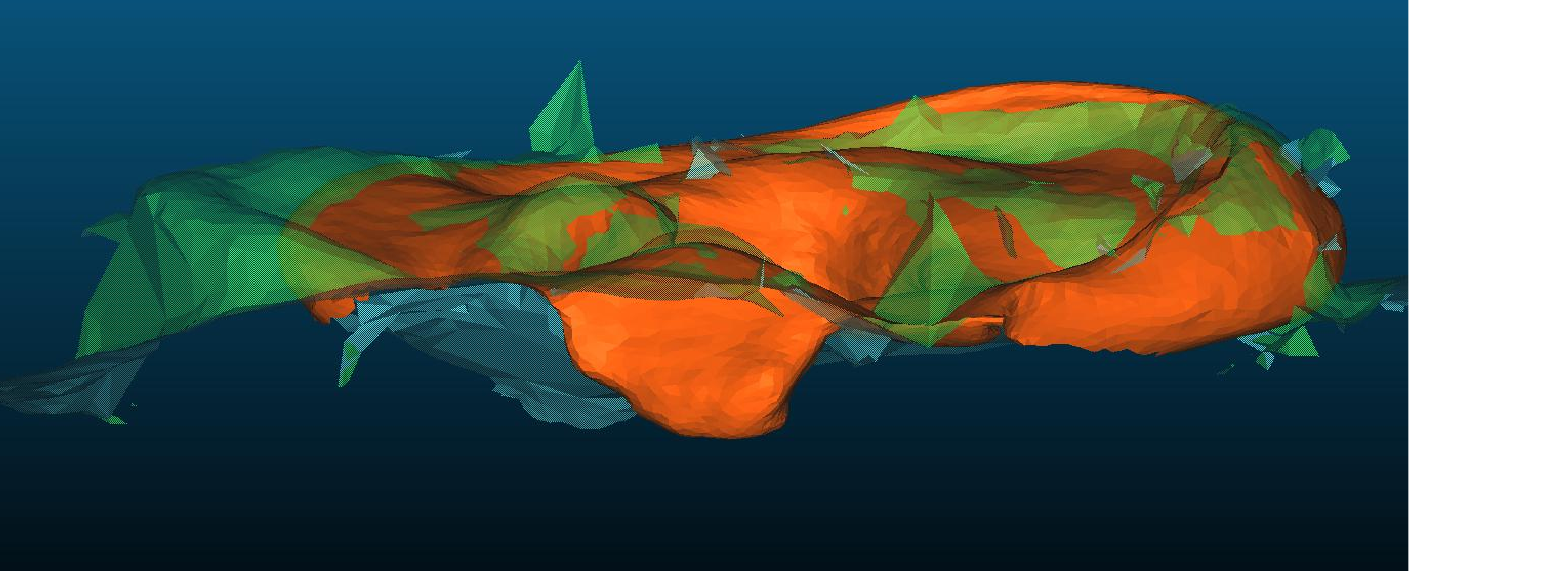}}
	
	\caption{Detail of shape completion step with PCA (on the left) and GP (on the right), for the same sample. The green shapes correspond to the original scan and the orange ones to the shape predicted by each approach. Looking at the small protuberance on the bottom of the ear, it is clear that the PCA model does not have enough variability to properly deform the template. On the other hand, the augmented kernel of the GP is able to provide the necessary variability, when given the same correspondences as the PCA model.}
	\label{fig:detail_shape_completion}
\end{figure*}

As expected, the PCA approach guarantees to produce a smooth ear shape, but the variability in the model is not enough to correctly complete the shape when the latter is too different from the dataset samples. The deformed template from the registration solves this problem in part but, since it has no further information on the shape, suffers too much deformation. The GP framework is a good compromise between the two previous options, as it includes a prior on shape, while the augmentation through the Gaussian kernel provides additional flexibility. Consequently, unlike the PCA approach, it is able to reach deformations not previously seen on the training dataset, such as the detail presented in Fig.~\ref{fig:detail_shape_completion}.

However, there are still shortcomings on the final shape, as in some parts it is not adequately fitted to the target. This suggests that further work can be done in finding a more appropriate kernel or including further steps on either the registration or shape completion steps. 

\begin{figure*}[htp]
	\centering
	\subfloat[]{
		\includegraphics[width=.35\textwidth,clip,  trim={150pt 20pt 150pt 50pt}]{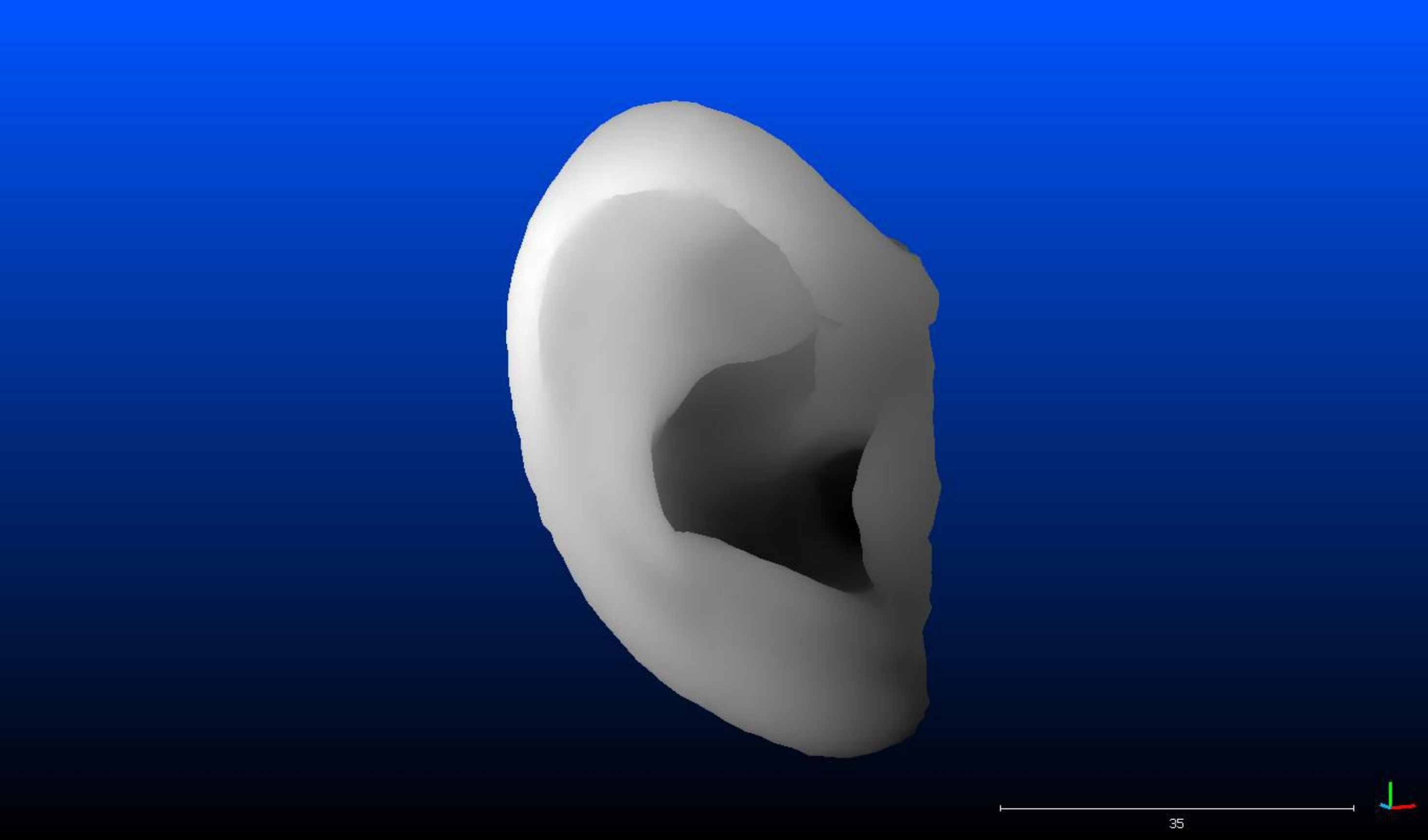}}
	~
	\subfloat[]{
		\includegraphics[width=.35\textwidth,clip,  trim={150pt 20pt 150pt 50pt}]{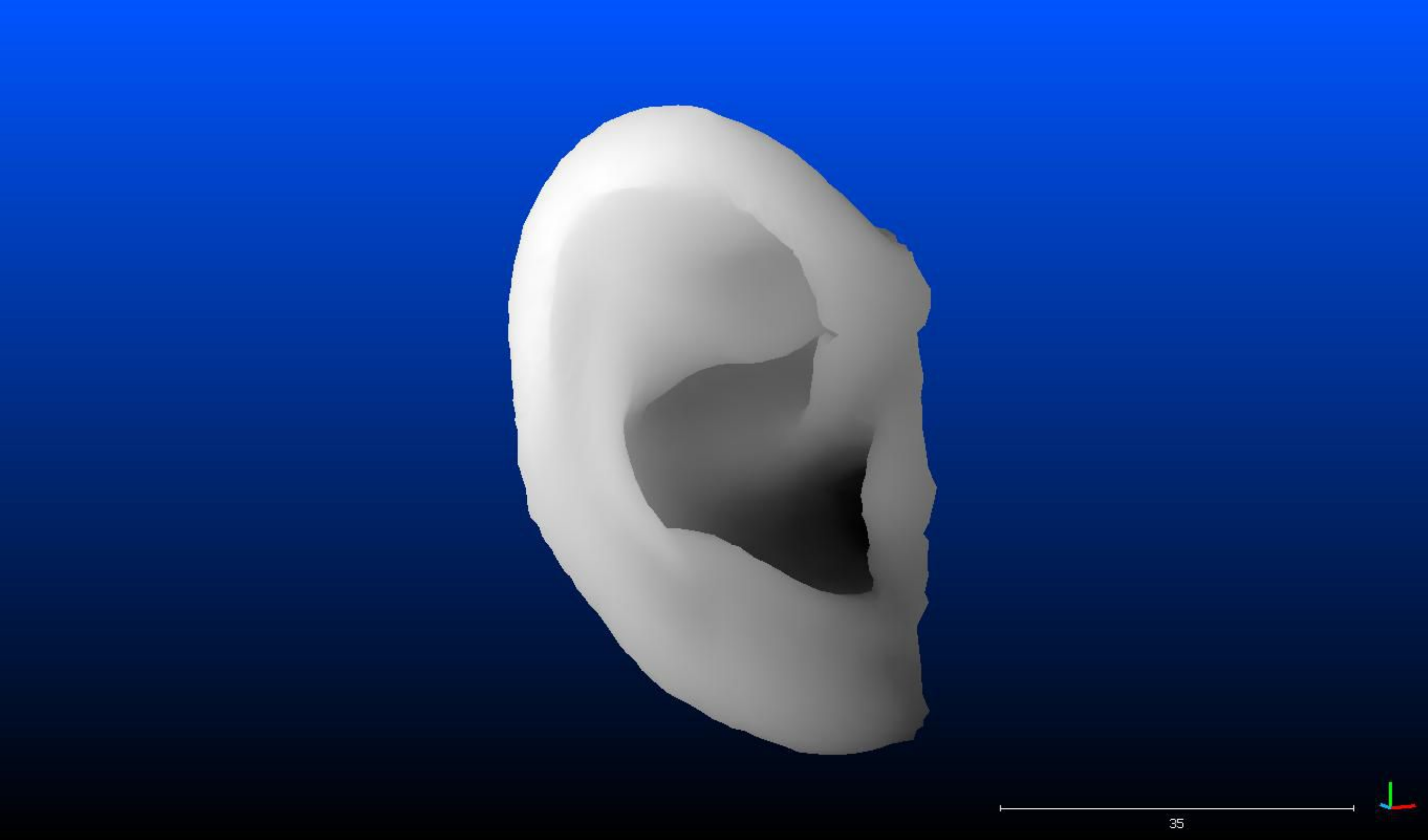}}
	
	\caption{Mean shape of Reconstructed Dataset with GP (left) and PPCA (right). }
	\label{fig:mean_shapes_reconstructed}
\end{figure*}

\begin{figure*}[htp]
	\subfloat[]{
		\includegraphics[width=.5\textwidth,clip]{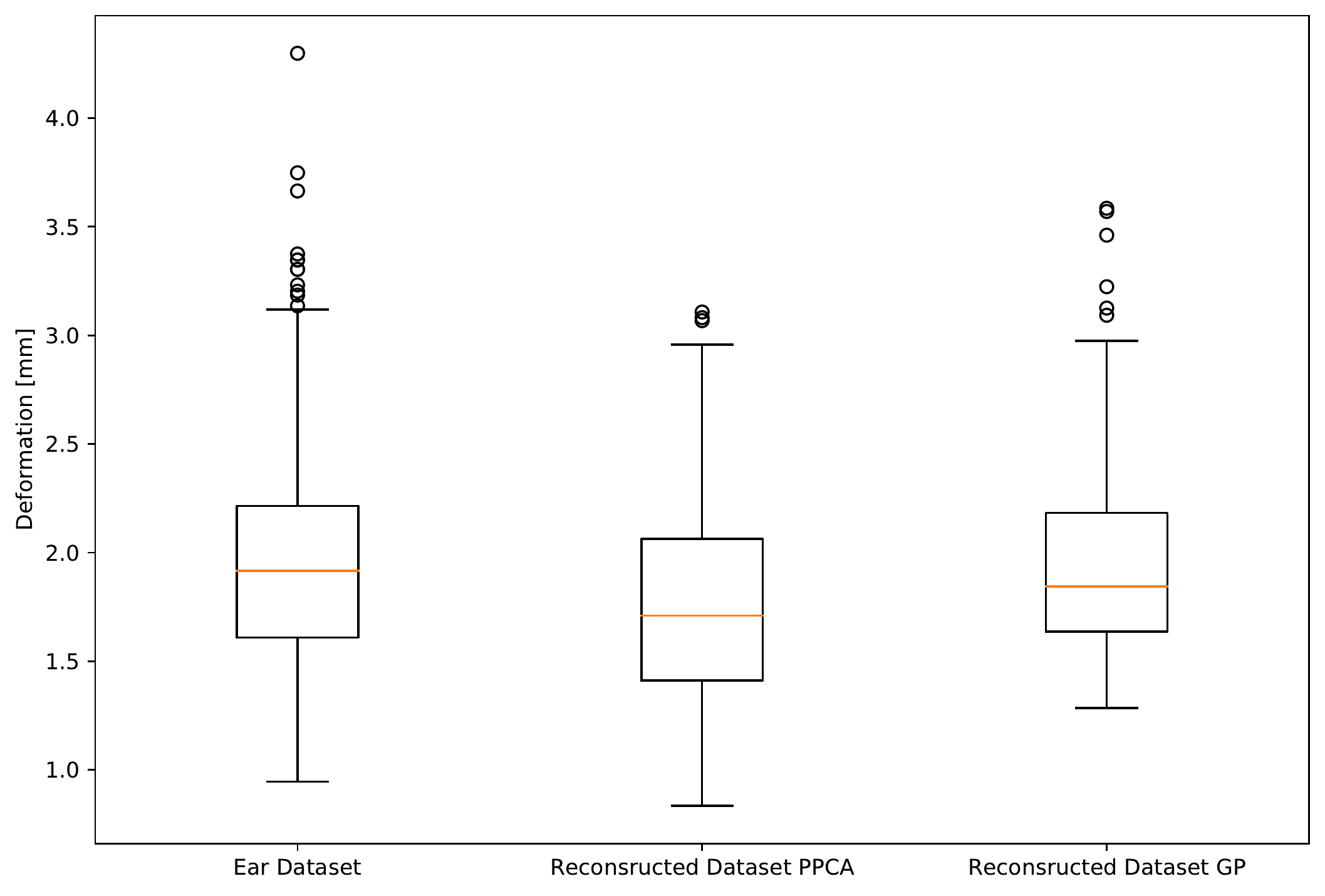}}
	~
	\subfloat[]{
		\includegraphics[width=.5\textwidth,clip]{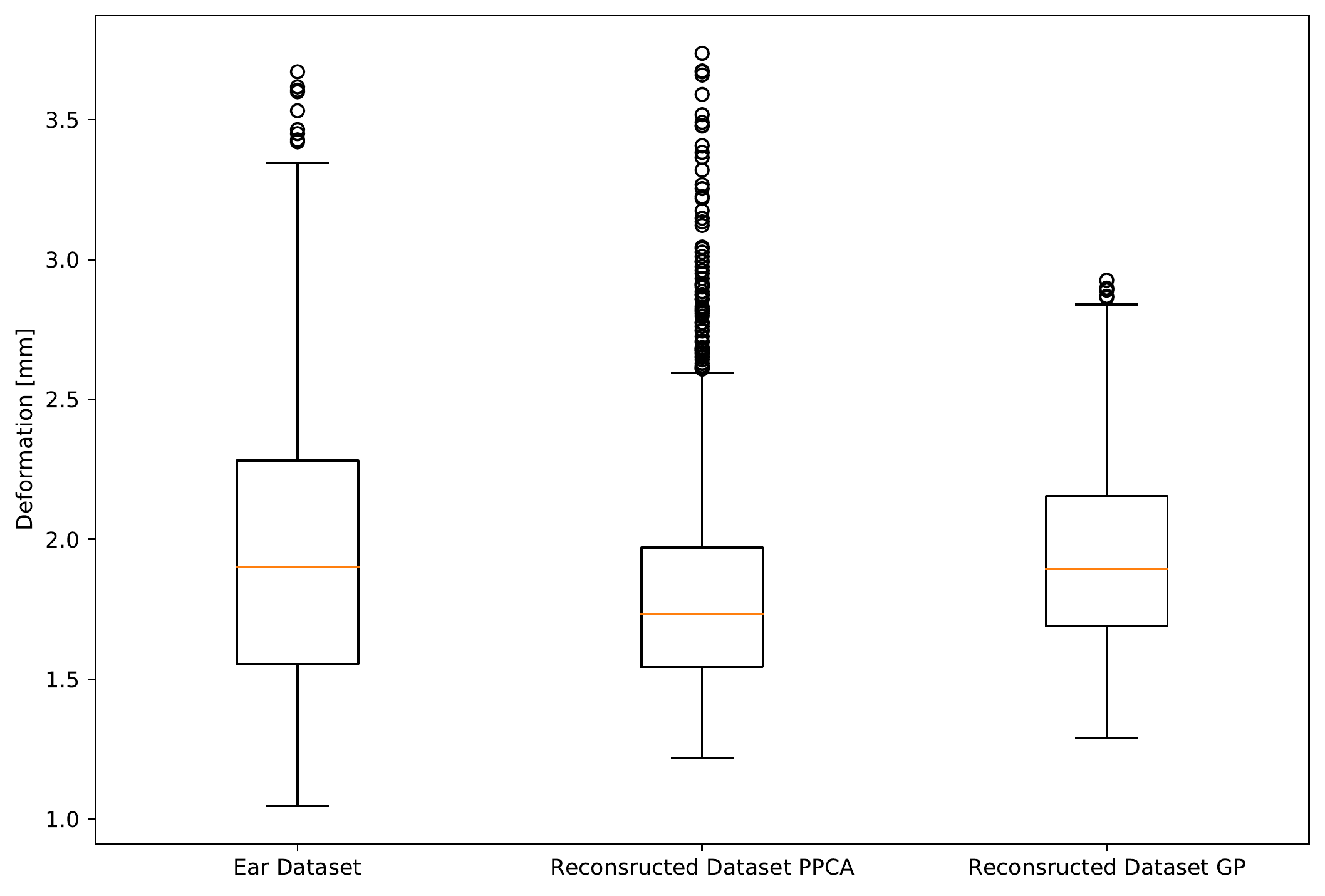}}
	
	\caption{Mean deformation of each point with respect to the mean shape, averaged over the shape (on the left) or the same point of all shapes (on the right). }
	\label{fig:boxplot_mean_deformation}
\end{figure*}

\begin{figure*}[htp]
	\subfloat[]{
		\includegraphics[width=\textwidth,clip]{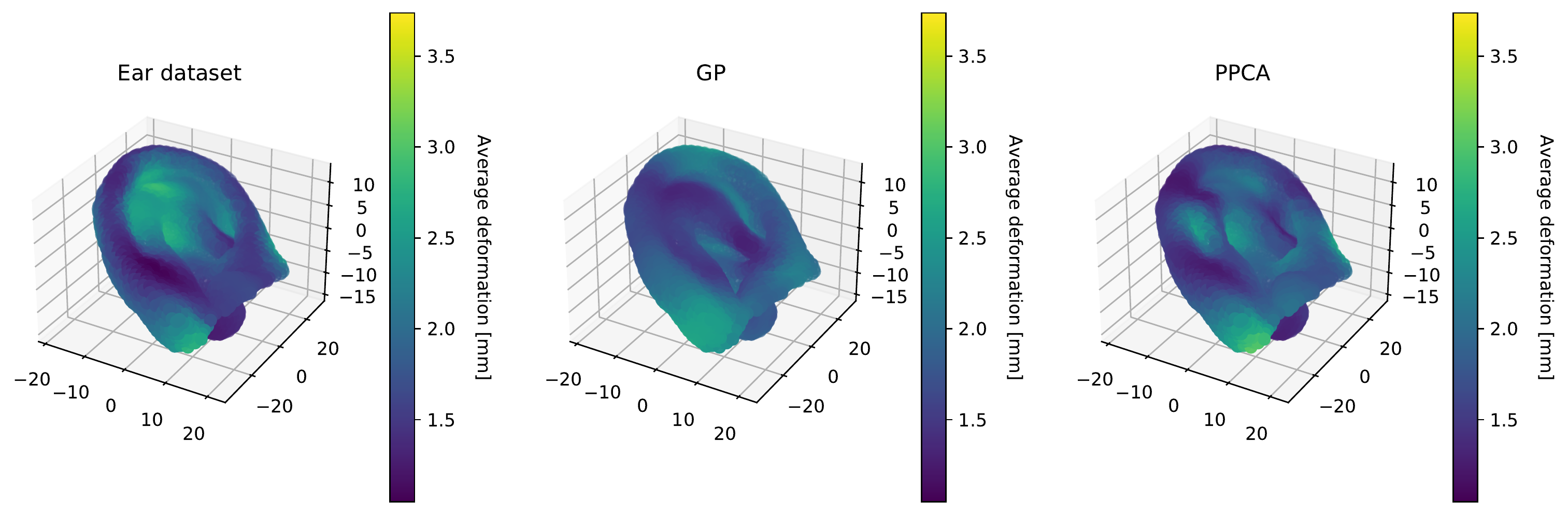}}
	\newline
	\subfloat[]{
		\includegraphics[width=\textwidth,clip]{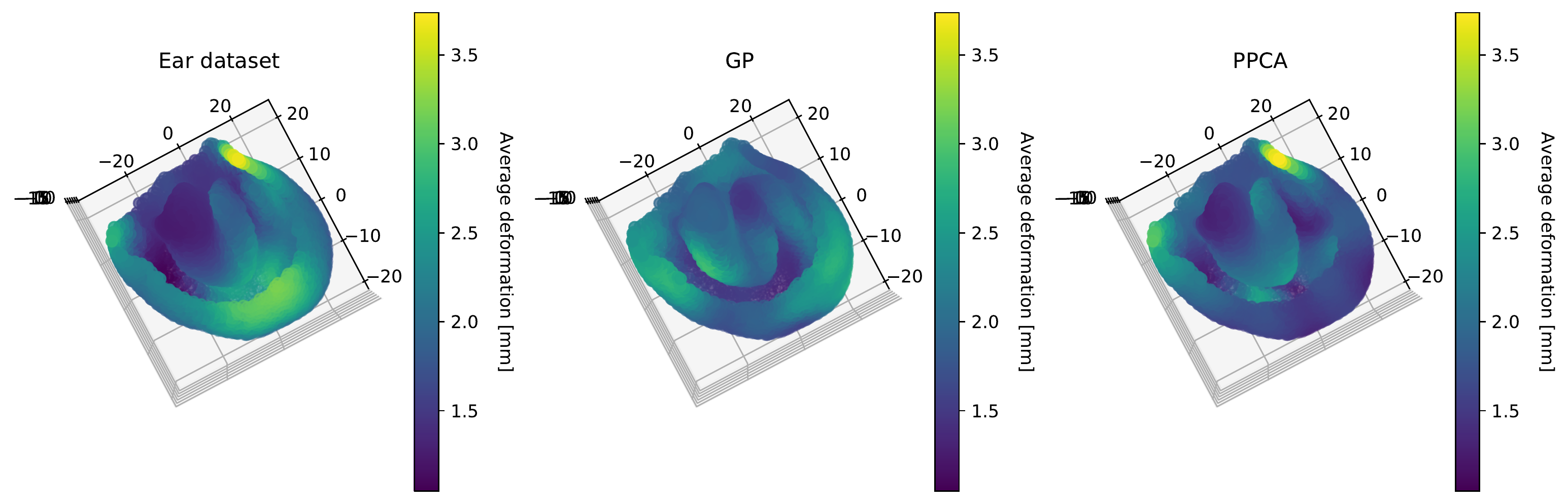}}
	
	\caption{Average deformation of each point (with respect to the mean shape) over all the shapes, for each dataset: Ear (on the left), Reconstructed with GP (middle) and Reconstructed with PPCA (right). The first row provides an upper view of the ear, while the second provides a bottom one, in order to facilitate visualization. Given that the PPCA approach is built on top of the Ear Dataset it is natural that both present a similar pattern of deformation. In contrast, GP does not have the same regions with the extremes of low and high deformation, having more moderate values over the entire shape.}
	\label{fig:plot3d}
\end{figure*}

After this visual comparison, we analyse the full produced datasets for both GP and PPCA shape completion. The primary goal is to compare both approaches and confront them with the equivalent metrics on the Ear dataset. We apply Generalized Procrustes Analysis (GPA) to all of the datasets in order to study only shape differences. After this, we compute the mean shape of each one (Figure~\ref{fig:mean_shapes_reconstructed}) and the deformation each point suffers with respect to it, i.e. the Euclidean distance between the correspondent point of the mean shape and each sample of the dataset (Figure~\ref{fig:boxplot_mean_deformation}). We observe the average deformation over the full shape and over the same point on each shape of the dataset, represented in the boxplots of Figure~\ref{fig:boxplot_mean_deformation}. We also represent the latter over a 3D plot of the mean shape of the Ear Dataset (Figure~\ref{fig:plot3d}), so that we can visualize where the deformations occur. Looking at the average shape deformation we see that  all the three datasets present similar values, with the Ear one having a slightly increased average. On the other hand, the GP one has less shapes similar to the mean one, while the PPCA one has a similar distribution with respect to the Ear, except it has less different shapes in general.

At this point, we may ask whether the overall shape differences in the Reconstructed Dataset are caused by a particular region which is not correctly fitted or registered, since we have observed that this is still a limitation of the pipeline. Observing the average deformation over point (Figure~\ref{fig:boxplot_mean_deformation} on the right), we observe that this is not the case given that even the regions with lower average deformation have an acceptable value and are not being compensated by regions of increased one. Looking at Figure~\ref{fig:plot3d}, while the PPCA and Ear Dataset have a small region of very high deformation, this is not present in the GP, indicating that possibly we need to allow for more deformation on the GP prior. On the other hand, we see that the latter has higher larger deformations on a large region of the bottom of the ear. This is a positive outcome, as one of the main obstacles we have found was the fitting of such area. 

We also look at the most different shapes of the Reconstructed Datasets on Figure~\ref{fig:mean_shapes_reconstructed}, that is the shape with higher average deformation with respect to the mean shape. As expected, while the PPCA has no unexpected deformations, the GP has a problematic upper area. This is observed rather frequently and calls for an improvement on the registration procedure with GP, since the problem lies in the fitting of the posterior part of the ear to the skin, incorrectly identified as inlier. This also means that part of the variability observed in the region on Figure~\ref{fig:plot3d} is due to the incorrect registration and will decrease. However, an improved process will allow for more deformation of the template, leading to complete fitting to the target shape, which is still a limitation, as seen in Figure~\ref{fig:real_pipeline}.

\begin{figure*}[htp]
	\centering
	\subfloat[]{
		\includegraphics[width=.4\textwidth,clip,  scale=0.8,trim={150pt 20pt 150pt 50pt}]{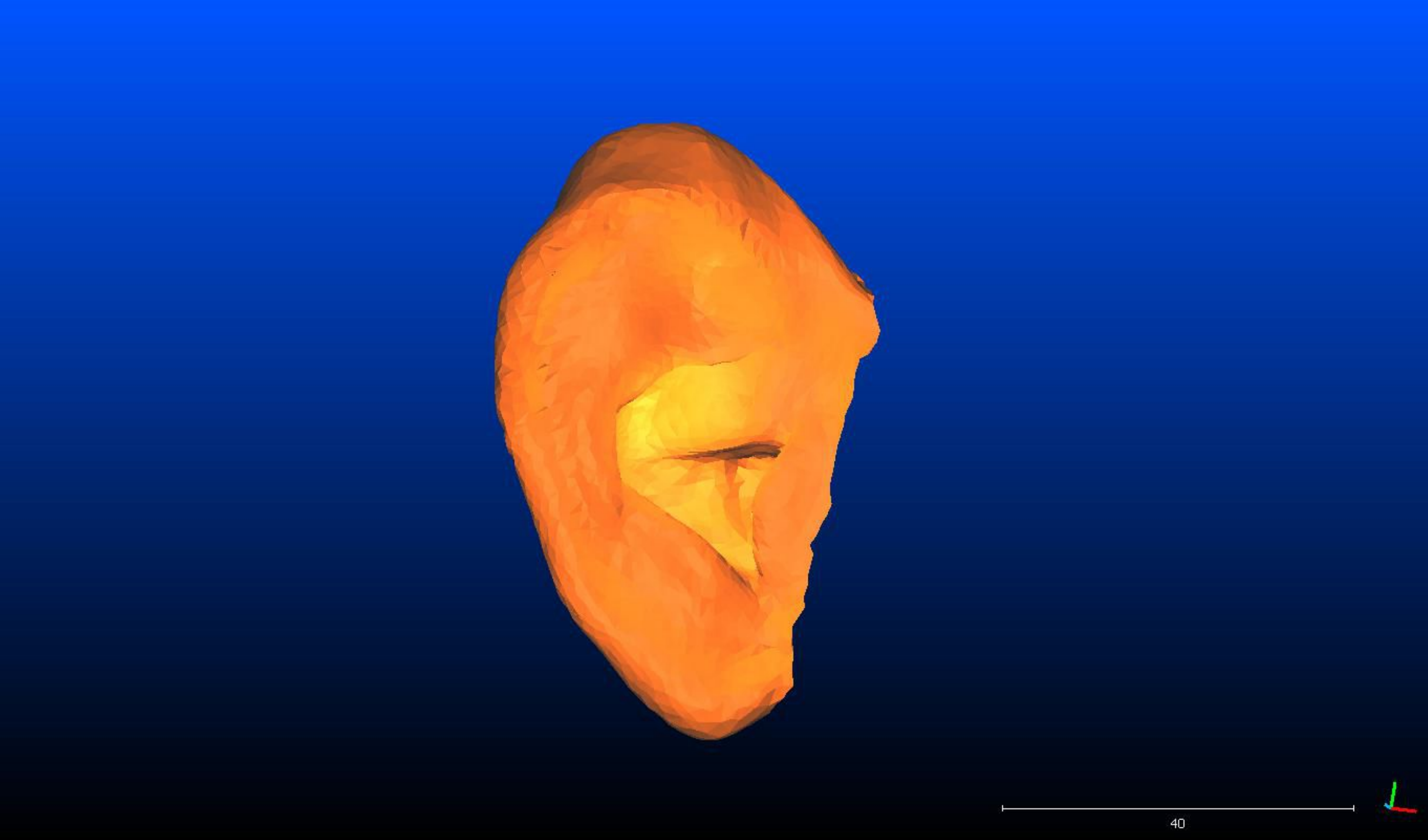}}
	~
	\subfloat[]{
		\includegraphics[width=.4\textwidth,clip,  trim={150pt 20pt 150pt 50pt}]{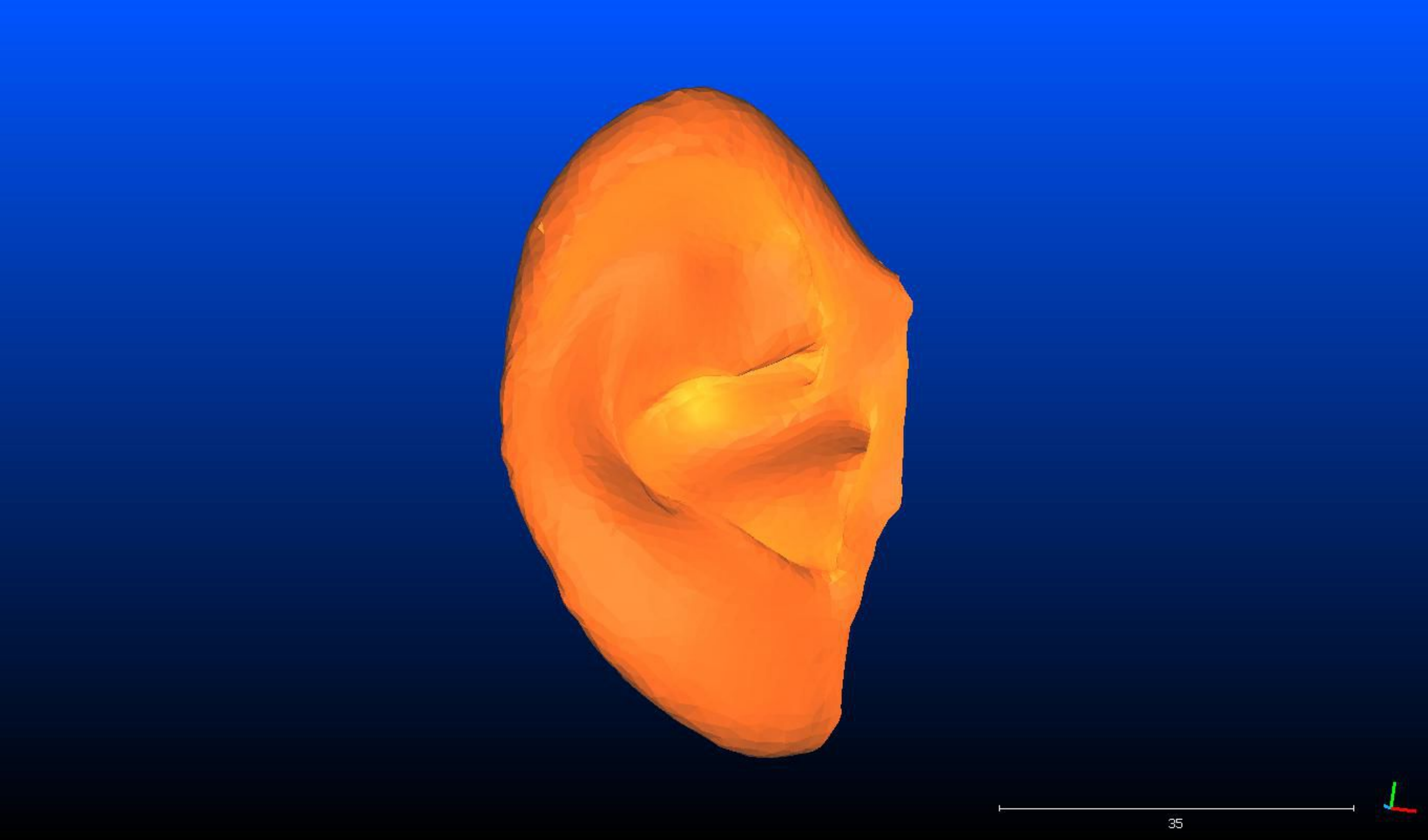}}
	
	\caption{Most different shape in the Reconstructed Dataset with GP (left) and PPCA (right). From each dataset we compute the average Euclidean distance with respect to its mean shape, in order to access how reasonable are the completed shapes. While the PPCA result has no unexpected deformations, the GP sample reflects the problems encountered in the registration on the top part. Nonetheless, it is still a reasonable result and does not exclude GP as strong candidate for shape completion, taking into account that the model parameters were not tuned beforehand.}
	\label{fig:diff_shapes_reconstructed}
\end{figure*}

\section{Concluding Remarks}
\label{sec:concl}

Taking the previous results into account, we propose the final pipeline composed by an initial registration with RANSIP, a registration refinement with BCPD and GP regression for shape completion. The proposed approach is able to complete the ears from the dataset in \cite{article:3DMM_LYHM}, although it still lacks some accuracy in matching the final shape. As future work, the pipeline must be improved in order to avoid those failures, either by improving the registration or by considering a different kernel in the GP framework. From the obtained Head dataset with the complete ears, it is also possible to obtain a relationship between the head shape and the ear one. This can help in building an improved ear model, which in turn may increase the performance of the current pipeline.

\bibliographystyle{ieeetr}
\bibliography{bibliography_db}{}

\begin{thebibliography}{10}

\bibitem{article:3DMM_LYHM}
H.~{Dai}, N.~{Pears}, W.~{Smith}, and C.~{Duncan}, ``A {3D} morphable model of
  craniofacial shape and texture variation,'' in {\em 2017 IEEE International
  Conference on Computer Vision (ICCV)}, Oct 2017.

\bibitem{article:3DMM_large_scale}
J.~{Booth}, A.~{Roussos}, S.~{Zafeiriou}, A.~{Ponniahy}, and D.~{Dunaway}, ``A
  {3D} morphable model learnt from 10,000 faces,'' in {\em 2016 IEEE Conference
  on Computer Vision and Pattern Recognition (CVPR)}, pp.~5543--5552, June
  2016.

\bibitem{article:Gaussian_Process_Combining_3DMM}
S.~Ploumpis, H.~Wang, N.~Pears, W.~A.~P. Smith, and S.~Zafeiriou, ``Combining
  {3D} morphable models: A large scale face-and-head model,'' in {\em 2019
  IEEE/CVF Conference on Computer Vision and Pattern Recognition (CVPR)}, 2019.

\bibitem{article:Gaussian_Process_Combining_3DMM_Ear}
S.~Ploumpis, E.~Ververas, E.~O. Sullivan, S.~Moschoglou, H.~Wang, N.~Pears,
  W.~A.~P. Smith, B.~Gecer, and S.~Zafeiriou, ``Towards a complete {3D}
  morphable model of the human head,'' {\em IEEE transactions on pattern
  analysis and machine intelligence}, 2020.

\bibitem{article:3DMM_Basel_Face_Model}
P.~{Paysan}, R.~{Knothe}, B.~{Amberg}, S.~{Romdhani}, and T.~{Vetter}, ``A {3D}
  face model for pose and illumination invariant face recognition,'' in {\em
  2009 Sixth IEEE International Conference on Advanced Video and Signal Based
  Surveillance}, pp.~296--301, Sep. 2009.

\bibitem{article:Morphable_Blanz_Vetter}
V.~Blanz and T.~Vetter, ``A morphable model for the synthesis of {3D} faces,''
  {\em SIGGRAPH'99 Proceedings of the 26th annual conference on Computer
  graphics and interactive techniques}, 09 2002.

\bibitem{article:3DMM_Ears_Dai_Data_augmented_journal_citation}
H.~Dai, N.~Pears, and W.~Smith, ``Augmenting a {3D} morphable model of the
  human head with high resolution ears,'' {\em Pattern Recognition Letters},
  vol.~128, pp.~378--384, 2019.

\bibitem{article:3DMM_Ears_Zolfaghari_LDDMM}
R.~{Zolfaghari}, N.~{Epain}, C.~T. {Jin}, J.~{Glaunès}, and A.~{Tew},
  ``Generating a morphable model of ears,'' in {\em 2016 IEEE International
  Conference on Acoustics, Speech and Signal Processing (ICASSP)},
  pp.~1771--1775, March 2016.

\bibitem{article:Ear_model_template_based}
Y.~Chu, K.~Zhang, H.~Wei, and Y.~Wang, ``Template-based ear modeling and
  reconstruction,'' in {\em 2019 Chinese Automation Congress (CAC)},
  pp.~4118--4123, 2019.

\bibitem{article:Registration_NICP_2007}
B.~{Amberg}, S.~{Romdhani}, and T.~{Vetter}, ``Optimal step nonrigid {ICP}
  algorithms for surface registration,'' in {\em 2007 IEEE Conference on
  Computer Vision and Pattern Recognition}, pp.~1--8, June 2007.

\bibitem{article:Registration_CPD}
A.~{Myronenko} and X.~{Song}, ``Point set registration: Coherent point drift,''
  {\em IEEE Transactions on Pattern Analysis and Machine Intelligence},
  vol.~32, pp.~2262--2275, Dec 2010.

\bibitem{database:SYMARE}
C.~T. {Jin}, P.~{Guillon}, N.~{Epain}, R.~{Zolfaghari}, A.~{van Schaik}, A.~I.
  {Tew}, C.~{Hetherington}, and J.~{Thorpe}, ``Creating the {S}ydney {Y}ork
  morphological and acoustic recordings of ears database,'' {\em IEEE
  Transactions on Multimedia}, vol.~16, no.~1, pp.~37--46, 2014.

\bibitem{article:Registration_ICP}
P.~J. {Besl} and N.~D. {McKay}, ``A method for registration of {3D} shapes,''
  {\em IEEE Transactions on Pattern Analysis and Machine Intelligence},
  vol.~14, no.~2, pp.~239--256, 1992.

\bibitem{article:Registration_ICP_Statistical}
S.~Cheng, I.~Marras, S.~Zafeiriou, and M.~Pantic, ``Statistical non-rigid {ICP}
  algorithm and its application to {3D} face alignment,'' {\em Image and Vision
  Computing}, vol.~58, 11 2016.

\bibitem{article:GO_ICP}
J.~Yang, H.~Li, D.~Campbell, and Y.~Jia, ``Go-{ICP}: A globally optimal
  solution to {3D} {ICP} point-set registration,'' {\em IEEE Transactions on
  Pattern Analysis and Machine Intelligence}, vol.~38, p.~2241–2254, Nov
  2016.

\bibitem{article:SDRSAC}
H.~Le, T.-T. Do, T.~Hoang, and N.-M. Cheung, ``{SDRSAC}: Semidefinite-based
  randomized approach for robust point cloud registration without
  correspondences,'' {\em Proceedings of the IEEE/CVF Conference on Computer
  Vision and Pattern Recognition}, 2019.

\bibitem{article:CPD_Extended_correspondence_priors}
V.~Golyanik, B.~Taetz, G.~Reis, and D.~Stricker, ``Extended coherent point
  drift algorithm with correspondence priors and optimal subsampling,'' 03
  2016.

\bibitem{article:CPD_unified}
G.~{Yang}, R.~{Li}, Y.~{Liu}, and J.~{Wang}, ``A unified framework for nonrigid
  point set registration via coregularized least squares,'' {\em IEEE Access},
  vol.~8, pp.~130263--130280, 2020.

\bibitem{article:CPD_PreservingGlobal_LocalStructures}
J.~{Ma}, J.~{Zhao}, and A.~L. {Yuille}, ``Non-rigid point set registration by
  preserving global and local structures,'' {\em IEEE Transactions on Image
  Processing}, vol.~25, no.~1, pp.~53--64, 2016.

\bibitem{article:CPD_adaptive_template_Dai_Pears}
H.~Dai, N.~Pears, and W.~Smith, ``Non-rigid {3D} shape registration using an
  adaptive template,'' pp.~48--63, 01 2019.

\bibitem{article:CPD_membership}
L.~{Fang}, Z.~{Sun}, and K.~{Lam}, ``An effective membership probability
  representation for point set registration,'' {\em IEEE Access}, vol.~8,
  pp.~9347--9357, 2020.

\bibitem{article:Registration_CPD_LocalConnectivity}
L.~{Bai}, X.~{Yang}, and H.~{Gao}, ``Nonrigid point set registration by
  preserving local connectivity,'' {\em IEEE Transactions on Cybernetics},
  vol.~48, no.~3, pp.~826--835, 2018.

\bibitem{article:Registration_CPD_Bayesian}
O.~{Hirose}, ``A {B}ayesian formulation of coherent point drift,'' {\em IEEE
  Transactions on Pattern Analysis and Machine Intelligence}, pp.~1--1, 2020.

\bibitem{article:Registration_Graph_Matching_Spectral_relaxation}
T.~Cour, P.~Srinivasan, and J.~Shi, ``Balanced graph matching,'' in {\em
  Advances in Neural Information Processing Systems 19} (B.~Sch\"{o}lkopf,
  J.~C. Platt, and T.~Hoffman, eds.), pp.~313--320, MIT Press, 2007.

\bibitem{article:Registration_Graph_Matching_SDP_relaxation}
C.~Schellewald and C.~Schnörr, ``Probabilistic subgraph matching based on
  convex relaxation,'' pp.~171--186, 11 2005.

\bibitem{article:Registration_Graph_Matching_HighOrder_randomwalks}
J.~Lee, M.~Cho, and K.~Lee, ``Hyper-graph matching via reweighted random
  walks,'' {\em Proceedings of the IEEE Computer Society Conference on Computer
  Vision and Pattern Recognition}, pp.~1633--1640, 06 2011.

\bibitem{article:Registration_Graph_Matching_HighOrder_tensorblock}
Q.~Nguyen, A.~Gautier, and M.~Hein, ``A flexible tensor block coordinate ascent
  scheme for hypergraph matching,'' 04 2015.

\bibitem{article:Probabilistic_PCA}
M.~Lüthi, T.~Albrecht, and T.~Vetter, ``Probabilistic modeling and
  visualization of the flexibility in morphable models,'' in {\em Mathematics
  of Surfaces}, pp.~251--264, 09 2009.

\bibitem{article:Gaussian_Process_2017_Vetter}
M.~{Lüthi}, T.~{Gerig}, C.~{Jud}, and T.~{Vetter}, ``Gaussian process
  morphable models,'' {\em IEEE Transactions on Pattern Analysis and Machine
  Intelligence}, vol.~40, pp.~1860--1873, Aug 2018.

\bibitem{article:GP_framework_low_rank_app}
J.~D{\"o}lz, T.~Gerig, M.~L{\"u}thi, H.~Harbrecht, and T.~Vetter,
  ``Error-controlled model approximation for {G}aussian process morphable
  models,'' {\em Journal of Mathematical Imaging and Vision}, vol.~61,
  pp.~443--457, 2018.

\end{thebibliography}
\begin{IEEEbiography}[{\includegraphics[width=1in,height=1.25in,clip,keepaspectratio]{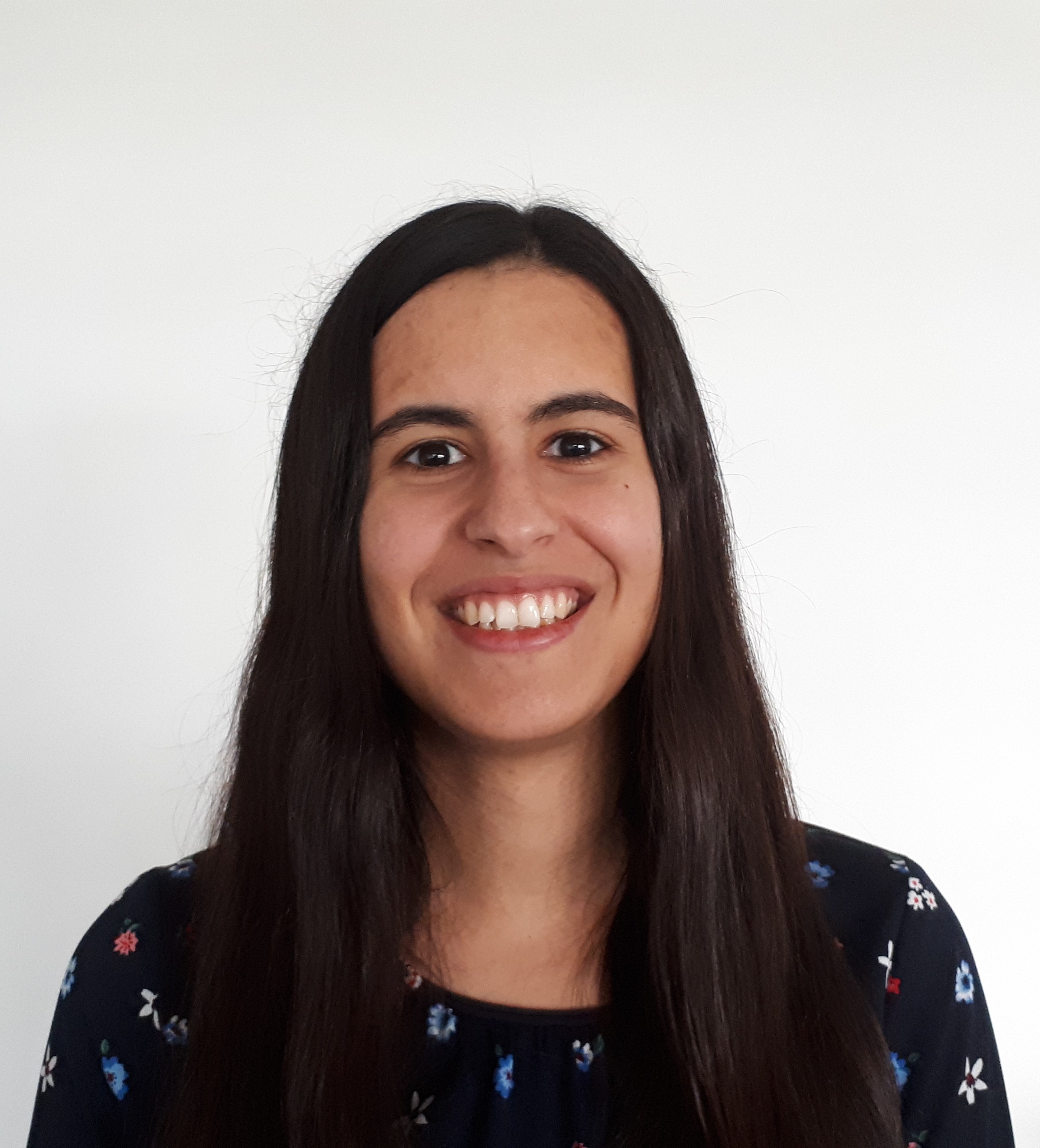}}]{Filipa Valdeira} 
has a BSc and MSc in Aerospcace Engineering from IST, Lisbon University, Portugal. Currently, she is a PhD student at the department of Mathematics, Università degli Studi di Milano, Italy, as part of the H2020 MSCA project BIGMATH. Her thesis focuses on Statistical Shape Analysis for 3D data and application of statistics and optimization for Big Data.
\end{IEEEbiography}
\begin{IEEEbiography}[{\includegraphics[width=1in,height=1.25in,clip,keepaspectratio]{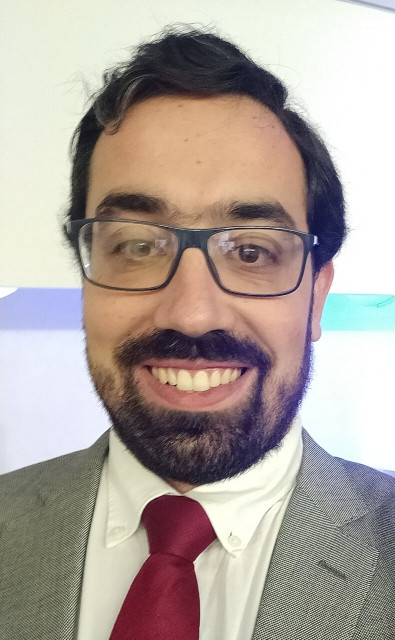}}]{Ricardo Ferreira} has a PhD degree (2010) in Electrotechnical Engineering from Instituto Superior Técnico (IST), Universidade Técnica de Lisboa. He was a researcher at Instituto de Sistemas e Robótica (ISR), Lisbon, in topics focused in solving geometrical problems arising in the areas of robotics, optimization and computer vision, where he coordinated (as co-PI) the Portuguese group involved in the European research project “RoboSoM” (EU-FP7-ICT-248366). He was also involved in the European research project “Poeticon++” (EU-FP7-ICT-288382) and in several national research and infrastructure projects from the Portuguese Foundation for Science and Technology (FCT), e.g. RBCog-Lab (Roteiro/0308/2013), AHA (CMUP-ERI/HCI/0046/2013), BIOMORPH (expl/eei-aut/2175/2013), DCCAL (PTDC/EEA-CRO/105413/2008). Since 2013 he is managing director of $\mu$Roboptics, where he is responsible for the development of several computer vision, robotics and process optimization projects for diverse applications and clients. He is currently responsible for the company participation in the FAST-bact research project (EU-H2020-FTI-730713) and several national research projects.
\end{IEEEbiography}
\begin{IEEEbiography}[{\includegraphics[width=1in,height=1.25in,clip,keepaspectratio]{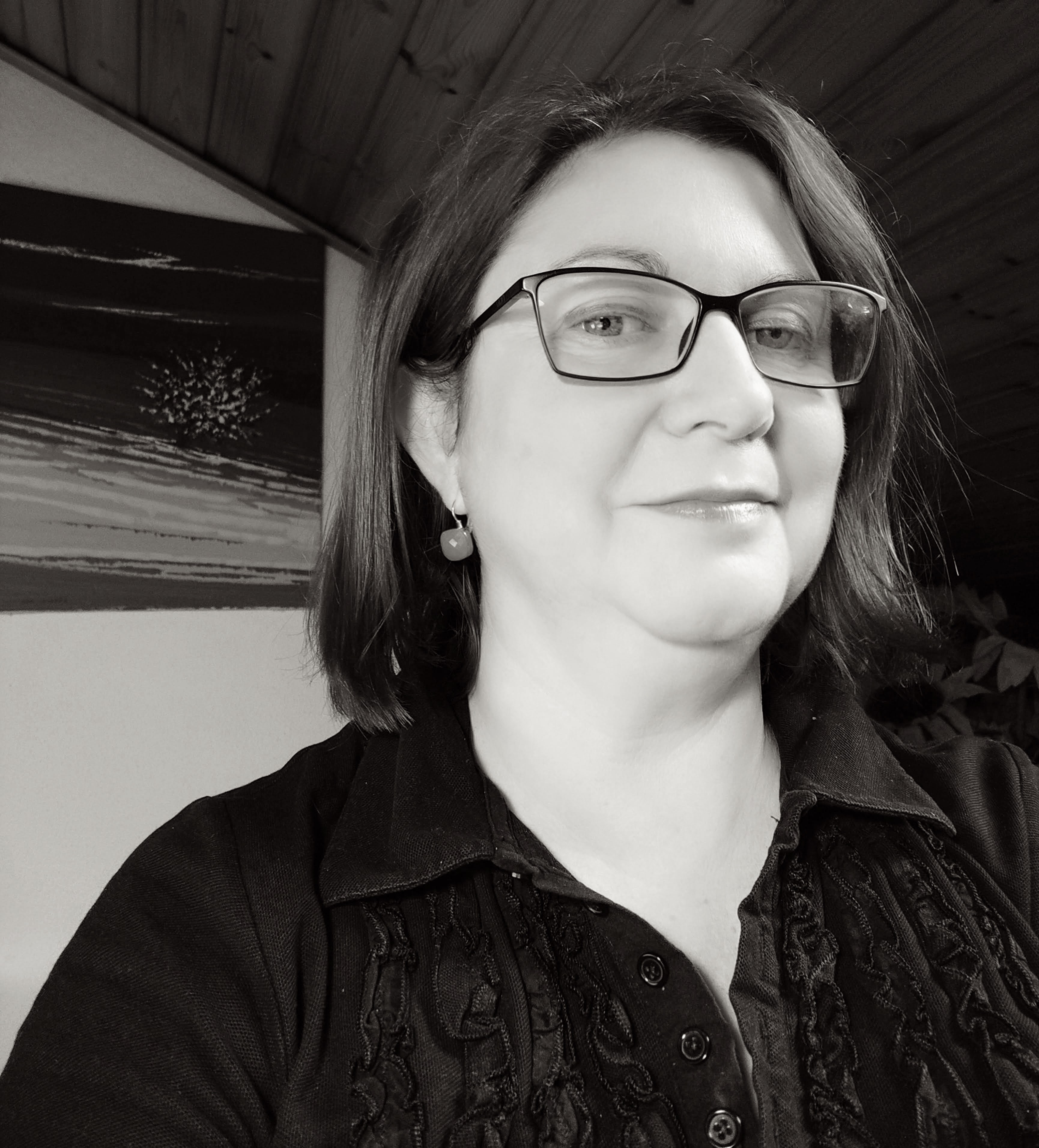}}]{Alessandra Micheletti } is an Associate Professor of Probability and Mathematical Statistics at the Dept. of Environmental Science and Policy, University of Milan. She got a PhD in Computational Mathematics and Operation Research in 1997. She is vice president of the European Consortium for Mathematics in Industry, a member of the Data Science Research Center of University of Milan and coordinator of the H2020 MSCA project BIGMATH. Her main research interests are in the field of Statistical Shape Analysis and Topological Data Analysis applied to industrial and life sciences problems.
\end{IEEEbiography}
\begin{IEEEbiography}[{\includegraphics[width=1in,height=1.25in,clip,keepaspectratio]{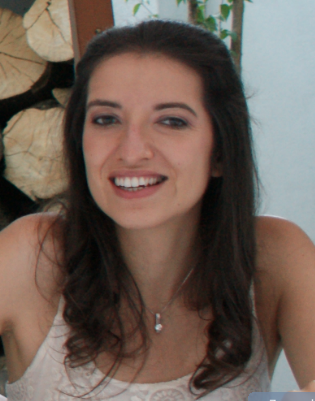}}]{Claudia Soares } is an Assistant Professor at NOVA School of Science and Technology, and a researcher at NOVA LINCS, Portugal. She uses real-world data problems to identify the shortcomings of current machine learning, data science, and Big Data methods. She applies optimization, statistics, and probability theory to address those gaps, developing robust, interpretable, and fair learning methods that can be trusted in real life. Her application areas are in healthcare, transportation, environmental and urban sciences, and space. Prof. Soares holds a degree in modern languages and literature, and a B.Sc., M.Sc., and Ph.D. in Electrical and Computer Engineering from Instituto Superior Técnico, Portugal. Prof. Soares is the national PI of the H2020 MSCA Innovative Training Network BIGMATH – Big Data Challenges for Mathematics, PI of a nationally-funded Data Science and AI for the public administration project in the area of healthcare management, and has many collaborations with industries like Epic Games, TAP Air Portugal, EDP, GMV, CUF Health, and tech SMEs like Neuraspace and uRoboptics. Her research is published in top journals in Signal Processing, Data Science, and data analytics.
\end{IEEEbiography}
\EOD

\end{document}